\documentclass[twoside,11pt]{article}
\title{Analytically Tractable Hidden-States Inference\\ in Bayesian Neural Networks}
\author{ \textsc{Luong Ha~Nguyen}\,\footnote{Equal contribution}\,\,\,$^{,}$\footnote{Correspondence: luongha.nguyen@gmail.com, james.goulet@polymtl.ca}  ~~and \textsc{James-A.~Goulet}\,\footnotemark[1]\,\,\,$^{,}$\footnotemark[2]
\mbox{}\\ 
\small Department of Civil, Geologic and Mining Engineering\\
\small \textsc{Polytechnique Montréal}, \underline{CANADA}\\}
\date{}

\usepackage{hyperref}       
\usepackage{url}            
\usepackage{booktabs}       
\usepackage{amsfonts,amssymb,latexsym,amscd,mathtools}       
\usepackage{nicefrac}       
\usepackage{microtype}      
\usepackage{bm}
\usepackage{multirow}
\usepackage{caption}
\usepackage{subfig}
\usepackage{graphicx}
\usepackage{todonotes}
\usepackage[T1]{fontenc}
\usepackage[ttdefault=true]{AnonymousPro}
\usepackage{soul}

\usepackage{cancel}
\usepackage[algoruled,linesnumbered,vlined,titlenotnumbered]{algorithm2e}
\newcommand\independent{\protect\mathpalette{\protect\independenT}{\perp}}
\def\independenT#1#2{\mathrel{\rlap{$#1#2$}\mkern2mu{#1#2}}}
\begin{document}

\maketitle

\begin{abstract}
With few exceptions, neural networks have been relying on backpropagation and gradient descent as the inference engine in order to learn the model parameters, because the closed-form Bayesian inference for neural networks has been considered to be intractable. In this paper, we show how we can leverage the \emph{tractable approximate Gaussian inference}'s (TAGI) capabilities to infer hidden states, rather than only using it for inferring the network's parameters. One novel aspect it allows is to infer hidden states through the imposition of constraints designed to achieve specific objectives, as illustrated through three examples: (1) the generation of adversarial-attack examples, (2) the usage of a neural network as a black-box optimization method, and (3) the application of inference on continuous-action reinforcement learning. These applications showcase how tasks that were previously reserved to gradient-based optimization approaches can now be approached with analytically tractable inference. \end{abstract}

\section{Introduction}\label{S:INTRO}
With few exceptions, neural networks have been relying on backpropagation \cite{rumelhart1988learning} and gradient descent as the inference engine in order to learn the model parameters. In such a case, the inference can be seen as approximating the posterior by a point solution minimizing a loss function. In addition to learning the model parameters, one may be interested in inferring the values of hidden states in a neural network. Note that we are not interested here in cases such as variational auto-encoder \cite{kingma2013auto}, or in generative adversarial networks  \cite{goodfellow2014generative,chen2016infogan} where dedicated latent variables are added; we are rather interested in inferring the value of hidden states from single observation instances. A first example is the case of adversarial attacks (AA), where images can be tailored in order to fool a neural network into performing incorrect classifications with high certainty \cite{43405}. In the context of white-box AA, images that seem realistic for a human observer, are generated by inferring perturbations that can be added to the input layer of a neural network in order to fool it. A second example of hidden state inference involves the definition of policy networks in reinforcement learning (RL) with methods such as advantage actor critic (A2C) \cite{mnih2016asynchronous} and 
proximal policy optimization (PPO) \cite{schulman2017proximal}. For such cases, current methods relying on backpropagation use gradient ascent in order to infer the optimal actions that are maximizing an action-value function \cite{sutton2011reinforcement}.

The closed-form Bayesian inference for neural networks has long been considered to be intractable, both in terms of its parameters \cite{goodfellow2016deep} or hidden states \cite{ardizzone2018analyzing,kruse2021benchmarking}. Recently, the \emph{tractable approximate Gaussian inference} (TAGI) \cite{goulet2020tractable} method was shown to either match or exceed the performance of neural networks trained with backpropagation in fully connected architectures \cite{goulet2020tractable}, for convolutional (CNN) and generative ones \cite{nguyen2021analytically}, as well as for deep reinforcement learning with categorical actions \cite{ha2021analytically}. This paper shows how we can leverage TAGI's probabilistic inference capabilities  to infer hidden states, rather than only using it for inferring the network's parameters. One novel aspect introduced is the capacity to infer hidden states through the imposition of constraints designed to attain specific objectives, as illustrated in this paper through three examples: (1) the generation of adversarial-attack examples, (2) the usage of a neural network as a black-box optimization method, and (3) the application of inference on continuous-action reinforcement learning. The paper is organized such that before diving in the theory and examples for these applications in Sections \ref{S:adversarial}-\ref{S:TAGI_PN}, Section \ref{S:tagi} reviews the theory behind TAGI. 

\section{Tractable Approximate Gaussian Inference}\label{S:tagi}
The tractable approximate Gaussian inference method relies on a two step \emph{forward-backward} process. In the \emph{forward} process, the uncertainty from the input layer is propagated through the hidden layers along with the uncertainty associated with model parameters, i.e., the weights and biases. The forward propagation of uncertainty allows forming the joint prior knowledge between successive pairs of hidden layers as well as between hidden layers and the parameters directly connecting into it. This process involves two approximations: first, that the product of a Gaussian hidden unit and a Gaussian weight parameter is also Gaussian, and second that non-linear activation functions can be locally linearized at the expected value of the hidden unit. Previous applications on validation benchmarks have confirmed that these approximations still allow matching or exceeding state-of-the-art performance on a same architecture trained with gradient descent and backpropagation \cite{goulet2020tractable,nguyen2021analytically,ha2021analytically}. 

The \emph{backward} process corresponds to the inference step that is based on the Gaussian conditional equations. In order to maintain a linear computational complexity during inference, we take advantage of the inherent conditional independence between the hidden layers of a neural network. This enables performing the layer-wise inference from successive pairs of hidden layers and from hidden layers to the parameters that connect into it. 

In the experimental setups explored so far, the inference capacity of TAGI was employed to learn the neural network parameters whereas the updated knowledge regarding the hidden units is discarded each time new training observations become available. In the current setup, we are not only interested in using TAGI to infer the network's parameters, but also the hidden units at specific location within the network. The appeal of TAGI is that it can inherently do so, without requiring any modifications to its formulation. In the following subsections, we will present how the novel inference capacity from the TAGI method can be leveraged in order to provide new solutions to existing challenges such as adversarial-attack generation, black-box optimization, and continuous-action reinforcement learning.

\section{Adversarial Attack through Inference}\label{S:adversarial}
In the first example, we are interested in white-box adversarial attacks \cite{akhtar2018threat} where we have access to the network and its parameters. Current white-box attacks are typically formulated as an optimization problem where one uses gradient descent and backpropagation in order to find optimal perturbations to be applied on the input layer in order to fool the network into making wrongful classifications.  
  
With TAGI, the generation of adversarial-attack images can be done analytically, without relying on an optimization process. We start with the assumption that we have a pre-trained neural network; Then, from a deterministic target image of size $\mathtt{M}\times\mathtt{N}$, for which we want to obtain a corrupt label, we define the prior knowledge on the input layer by the mean vector corresponding to the deterministic image $\bm{\mu}_{\bm{X}}=\bm{x}\in \mathbb{R}^{\mathtt{M}\times\mathtt{N}}$, and a diagonal covariance $\bm{\Sigma}_{\bm{X}}=\sigma_X^2\cdot \bm{I}$. Here, the amount of change that TAGI will apply on the original image during the inference procedure is controlled by the input layer's standard deviation parameter $\sigma_X$. This prior knowledge about the target image is propagated forward through the network analogously to the procedure presented in \S\ref{S:tagi}. Then, when it is time to observe the label, the correct one is replaced by the target label $\tilde{y}$ that is chosen for the attack. After performing the inference step, the initial image defined by its updated mean vector $\bm{\mu}_{\bm{X}|\tilde{y}}$ and covariance $\bm{\Sigma}_{\bm{X}|\tilde{y}}$ is now modified in order to trigger the class $\tilde{y}$. In order to improve the quality of the attack, the process is repeated recursively over multiple iterations, where the inferred values $\{\bm{\mu}_{\bm{X}|\tilde{y}}^{(i)},
\mathbf{\Sigma}_{\bm{X}|\tilde{y}}^{(i)}\}$ at iteration $i$ are used as the prior's hyper-parameters at the next iteration $i+1$. 

\begin{figure}[hb]
\centering
\subfloat[][MNIST]{\includegraphics[width=52mm]{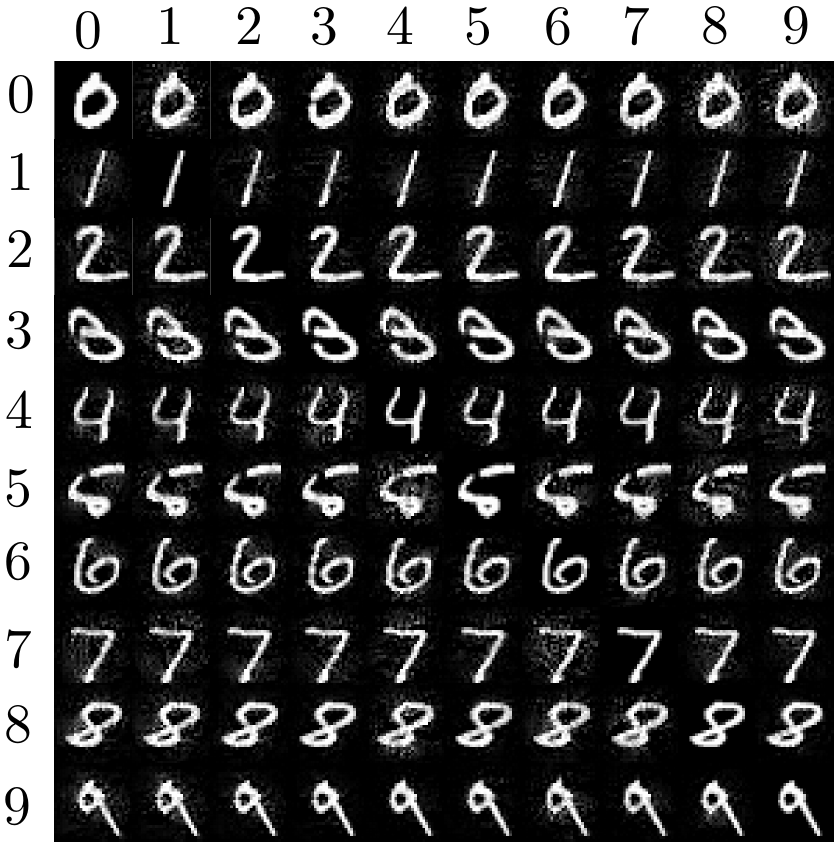}}~~\quad\subfloat[][Cifar10]{\includegraphics[width=52mm]{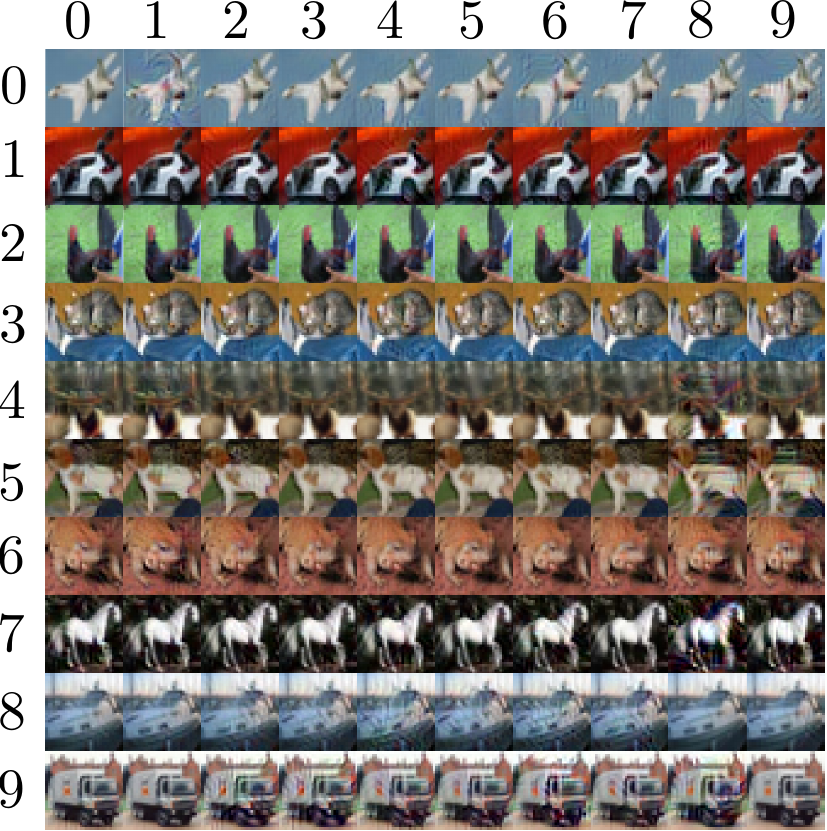}}
\caption{Examples of images subjected to adversarial attacks with different target labels $\tilde{y}$. Columns represent the different target labels $\tilde{y}$ and rows are the true label images.}
\label{FIG:AA}
\end{figure}

Figure \ref{FIG:AA} presents two examples where pre-trained convolutional neural networks \cite{nguyen2021analytically} are employed to generate attacks for the images from a) the MNIST and b) the Cifar10 dataset. For all experiments, we set $\sigma_{X} = 0.03$ with a maximal number of epochs $\mathtt{E} = 100$. The networks' details are presented in Appendix \ref{A:AAS}. 

Table \ref{TAB_MNIST} compares the error rates obtained: without attack, with targeted attacks where a specific class is seeked, and with non-targeted attacks where the goal is simply to fool the network. These results obtained for convolutional architectures confirm that TAGI can, without relying on an optimization scheme, infer adversarial-attack examples that are visually indistinguishable from the original.
\begin{table}[t]
\centering
\caption{Quantitative performance evaluation of TAGI's adversarial attacks on MNIST and Cifar10.}
\begin{tabular}{llc|c|c}
    	\addlinespace
    \toprule
    \multicolumn{2}{l}{}&\multicolumn{3}{c}{Error Rate [\%]}\\[2pt]
    \cmidrule{3-5}
    \multicolumn{1}{l}{Dataset}&\multicolumn{1}{l}{Model}&\multicolumn{1}{c}{No attack}&\multicolumn{1}{|c}{Targeted attack}&\multicolumn{1}{|c}{Non-targeted attack}\\[0pt] 
	\cmidrule{3-5}
\cmidrule{1-5}
\multirow{1}{*}{MNIST}&2 conv. &$0.64$&99.8&99.9\\[2pt]
Cifar10&3 conv.  &$22.0$&99.6&99.9\\[2pt]
\bottomrule
\end{tabular}
\label{TAB_MNIST}
\end{table}

\section{Optimization through Inference}\label{S:opt}
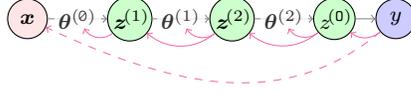
\begin{figure}[htbp]\centering
\tikzset{dist1/.style={path picture= {
    \begin{scope}[x=1pt,y=10pt]
      \draw plot[domain=-6:6] (\x,{1/(1 + exp(-\x))-0.5});
    \end{scope}
    }
  }
}
\tikzset{dist2/.style={path picture= {
    \begin{scope}[x=1pt,y=10pt]
      \draw plot[domain=-6:6] (\x,{\x/10});
    \end{scope}
    }
  }
}
\tikzstyle{input}=[draw,fill=red!10,circle,minimum size=15pt,inner sep=0pt]
\tikzstyle{hidden}=[draw,fill=green!20,circle,minimum size=15pt,inner sep=0pt]
\tikzstyle{output}=[draw,fill=blue!20,circle,minimum size=15pt,inner sep=0pt]
\tikzstyle{bias}=[draw=none,circle,minimum size=15pt,inner sep=0pt]

\tikzstyle{stateTransition}=[->,line width=0.25pt,draw=black!50]
\tikzstyle{Inference}=[->,line width=0.25pt,draw=magenta!75]

\begin{tikzpicture}[scale=1.36]\scriptsize
    \node (x)[input]   at (0, 0) {$\bm{x}$};
    \node (z1)[hidden] at (1, 0) {$\,\bm{z}^{\!(1)}$};
    \node (z2)[hidden] at (2, 0) {$\,\bm{z}^{\!(2)}$};
    \node (zO)[hidden] at (3,0) {$\,{z}^{\!(\mathtt{O})}$};

    \node (y)[output] at (3.6,0) {${y}$};
    
    \node (xp)[minimum size=7pt]   at (0.5, 0) {};
    \node (z1p)[minimum size=7pt] at (1.5, 0) {};
    \node (z2p)[minimum size=7pt] at (2.5, 0) {};
    \node (zLp)[minimum size=7pt] at (3.35, 0) {};

    \draw[stateTransition,opacity=1] (x) -- (z1) node [midway, rotate=0,fill=white,opacity=0.8] {\scriptsize$\!\bm{\theta}^{(\texttt{0})}\!$};
    \draw[stateTransition,opacity=1] (z1) -- (z2) node [midway, rotate=0,fill=white,opacity=0.8] {\scriptsize$\!\bm{\theta}^{(1)}\!$};
     \draw[stateTransition,opacity=1] (z2) -- (zO) node [midway, rotate=0,fill=white,opacity=0.8] {\scriptsize$\!\bm{\theta}^{(2)}\!$};
     \draw[stateTransition,opacity=1] (zO) -- (y);

      \draw[Inference,opacity=1] (z2) to [out=-140,in=-40] (z1);
      \draw[Inference,opacity=1] (zO) to [out=-140,in=-40] (z2);
      \draw[Inference,opacity=1] (y) to [out=-140,in=-40] (zO);
      
      \draw[Inference,opacity=1] (z1) to [out=-140,in=-60] (xp);
      \draw[Inference,opacity=1] (z2) to [out=-140,in=-60] (z1p);
      \draw[Inference,opacity=1] (zO) to [out=-140,in=-60] (z2p);

      \draw[Inference,opacity=1,dashed] (y) to [out=-140,in=-25] (x);

\end{tikzpicture}\vspace*{-4mm}
\caption{Compact representation of the variable nomenclature and the dependencies associated with a feedforward neural network with two hidden layers. }
\label{FIG:FNN}
\end{figure}

This section presents how we can leverage TAGI's inference capabilities to find the local maxima or minima of a function. In general, a feedforward neural network (FNN) is  a function approximation such that
\begin{equation}
y=g(\bm{x};\bm{\theta})+v,
\end{equation}
where $\bm{x}\in \mathbb{R}^\mathtt{X}$ is a vector of covariates, $y\in \mathbb{R}$ is the observed system response, $v\in \mathbb{R}$ is the observation error, and $\bm{\theta}$ is a vector of the parameters defining the weights $\bm{w}$ and biases $\bm{b}$ from the neural network $g(\bm{x};\bm{\theta})\in \mathbb{R}$. Figure \ref{FIG:FNN} presents a compact representation for the directed acyclic graph (DAG) describing the dependency between the different components of such a FNN. The red node describes the input layer, green nodes either hidden or output layers, the purple node is an observed system response, and gray arrows represent the dependencies encoded in the parameters of the network. The red arrows represent the flow of information during the inference procedure described in \S\ref{S:tagi}, where TAGI infers the weights and biases of the network in a layer-wise fashion in order to maximize the computational and memory efficiency. 

Once the parameters $\bm{\theta}$ of a neural network are learned, we can use the hidden units on the output layer ${z}^{(\mathtt{O})}=g(\bm{x};\bm{\theta})$ to predict the responses associated with covariates $\bm{x}$, and we also have access to  the $n^{\text{th}}$ derivatives of the function approximation,
\begin{equation}
g^n(\bm{x};\bm{\theta})=\frac{\partial^n {z}^{(\mathtt{O})}}{\partial\bm{x}^n}.
\end{equation} The details regarding the analytical calculation of partial derivatives using TAGI are presented in Appendix \ref{A:DTAGI}. In the context of an optimization problem,  the goal is to identify the input $\bm{x}$ that maximizes or minimizes ${z}^{(\mathtt{O})}$, at which the first derivative of the function approximation is equal to zero, i.e., $g^1(\bm{x};\bm{\theta})={z}^{'(\mathtt{O})}=0$. Using the same inference procedure presented in \S\ref{S:tagi}, we can infer analytically the probability density function (PDF) $f(\bm{x}|{z}^{'(\mathtt{O})}=0)=\mathcal{N}(\bm{x};\bm{\mu}_{\bm{X}|{z}^{'}},\bm{\Sigma}_{\bm{X}|{z}^{'}})$. For that purpose, we first define the prior knowledge for the vector of covariates $\bm{x}$ so that $\bm{X}\sim\mathcal{N}(\bm{x};\bm{\mu}_{\bm{X}},\bm{\Sigma}_{\bm{X}})$. Then, the expected value $\bm\mu_{\bm{X}|{z}^{'}}$ and variance $\bm{\Sigma}_{\bm{X}|{z}^{'}}$ are computed following
\begin{equation}
\label{eq_fwdtagi}
\begin{array}{rcl}
\bm\mu_{\bm{X}|{z}^{'}} &=& \bm\mu_{\bm{X}} - \bm\Sigma_{Z^{'(\mathtt{O})}\bm{X}}^{\intercal}\left({\sigma}_{{Z'}}^{(\mathtt{O})}\right)^{-2}\mu_{{Z'}}^{(\mathtt{O})}\\[8pt]
\bm\Sigma_{\bm{X}|{z}^{'}} &=&\bm\Sigma_{\bm{X}} - \bm\Sigma_{Z^{'(\mathtt{O})}\bm{X}}^{\intercal}\left({\sigma}_{{Z'}}^{(\mathtt{O})}\right)^{-2}\bm\Sigma_{Z^{'(\mathtt{O})}\bm{X}},
\end{array}
\end{equation}
where the expected value $\mathbb{E}[{Z}^{'(\mathtt{O})}]={\mu}_{{Z'}}^{(\mathtt{O})}$, variance $\text{var}[{Z}^{'(\mathtt{O})}]=({\sigma}_{{Z'}}^{(\mathtt{O})})^2$, and covariance $\mathbf{\Sigma}_{Z^{'(\mathtt{O})}\bm{X}}=\text{cov}({Z}^{'(\mathtt{O})},\bm{X})$ are obtained using the forward propagation of uncertainty defined for TAGI. In order to ensure that the inferred values for $\bm{x}$ correspond to either a minimum or a maximum, we need to rely the sign of the first derivative to control the direction of the mean update step. The expected value in Equation \ref{eq_fwdtagi} is thus reformulated as 
\begin{equation}
\label{eq_opttagi}
\bm{\mu}_{\bm{X}|{z}^{'}}=\bm{\mu}_{\bm{X}}+\alpha\cdot\text{sign}\!\left(\tfrac{\partial z^{'(\mathtt{O})}}{\partial \bm{x}}\right)\left|\mathbf{\Sigma}_{\mathtt{0}\bm{X}}^{\intercal}\left({\sigma}_{{Z'}}^{(\mathtt{O})}\right)^{-2}{\mu}_{{Z}}^{'(\mathtt{O})}\right|,
\end{equation}
where $\alpha=1$ when seeking a maximum, and $-1$ for a minimum.

In order to seek the location where the derivative is equal to zero, we repeat the inference multiple times where the inferred values $\{\bm{\mu}_{\bm{X}|{z}^{'}}^{(i)},
\mathbf{\Sigma}_{\bm{X}|{z}^{'}}^{(i)}\}$ at iteration $i$ are used as the prior's hyper-parameters at the next iteration $i+1$. The algorithm \ref{alg_opt} presents an example of the implementation for the optimization of a function using TAGI's inference capacity. 
\begin{figure}[h!]
\begin{center}
\begin{minipage}{1\linewidth}
\SetCustomAlgoRuledWidth{12.5cm}
\IncMargin{2pt}
\begin{algorithm}[H]
Define a neural network $g(\bm{x};{\bm\theta})$;\\[2pt]
Initialize $\sigma_{{V}}$, the prior for $\bm\theta$ and for the covariates $\bm{X}$; \\[2pt]
Given a dataset $\mathcal{D}=\{\bm{x}_{i}, y_{i}\}\,, \forall i=\{1, 2, \hdots,\mathtt{D}\}$;\\[2pt]
 \For{epoch $= 1 : \mathtt{E}$}{
\For{$i = 1 : \mathtt{D}$}{
Compute the prediction for a given input $\bm{x}_{i}$;\\[2pt]
$\left\{\mu_{Y}, \sigma_{Y}^2\right\}=g\left(\bm{x}_{i};\bm\theta\right)$;\\[2pt]
Update $\bm\mu_{\bm\theta|\mathcal{D}}$, $\bm\Sigma_{\bm\theta|\mathcal{D}}$ using TAGI;\\[2pt] 
Compute the partial derivative of $g\left(\bm{x};\bm\theta\right)$ w.r.t. $\bm{x}$;\\[2pt]
Update $\bm\mu_{\bm{X}|{z}^{'}}$, $\bm\Sigma_{\bm{X}|{z}^{'}}$ using Equation \ref{eq_opttagi};
 }
 }
\caption{Optimization of a function using TAGI}
\label{alg_opt}
\end{algorithm}
\end{minipage}
\end{center}
\end{figure}

\begin{figure}[htbp]
\centering
\,\subfloat[Function \& training data]{\includegraphics[width=60mm]{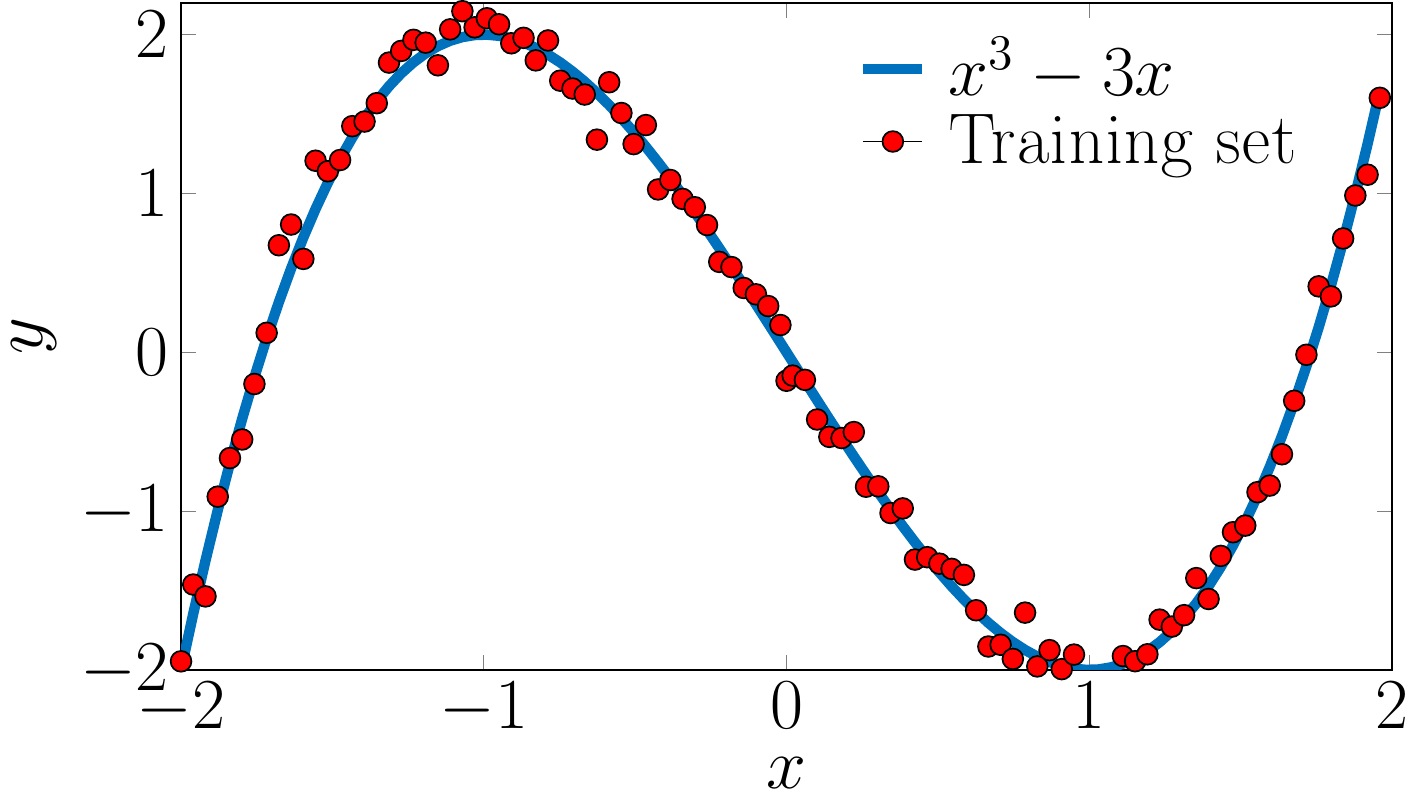}}~~~
\subfloat[TAGI function approximation]{\includegraphics[width=60mm]{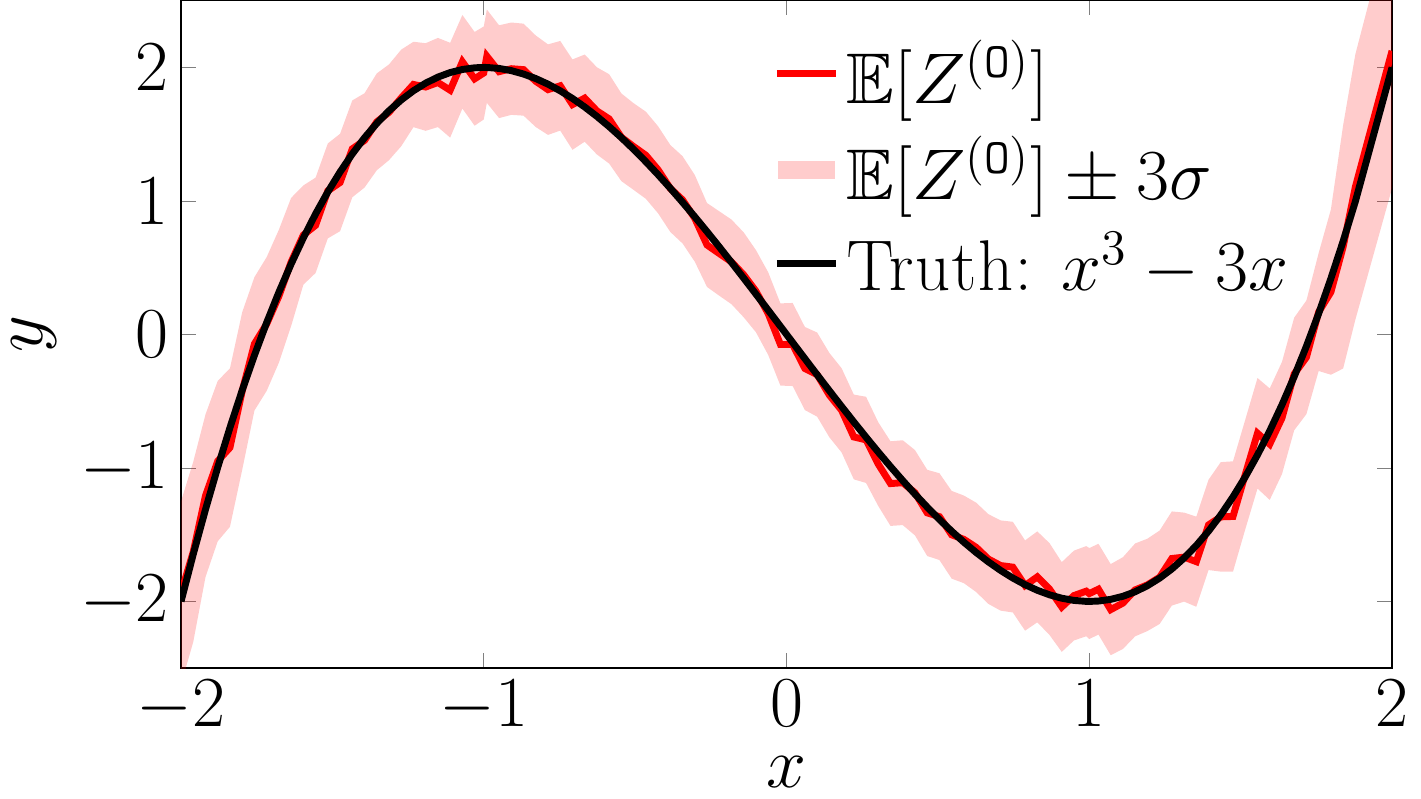}}\\\subfloat[TAGI derivative]{\includegraphics[width=60mm]{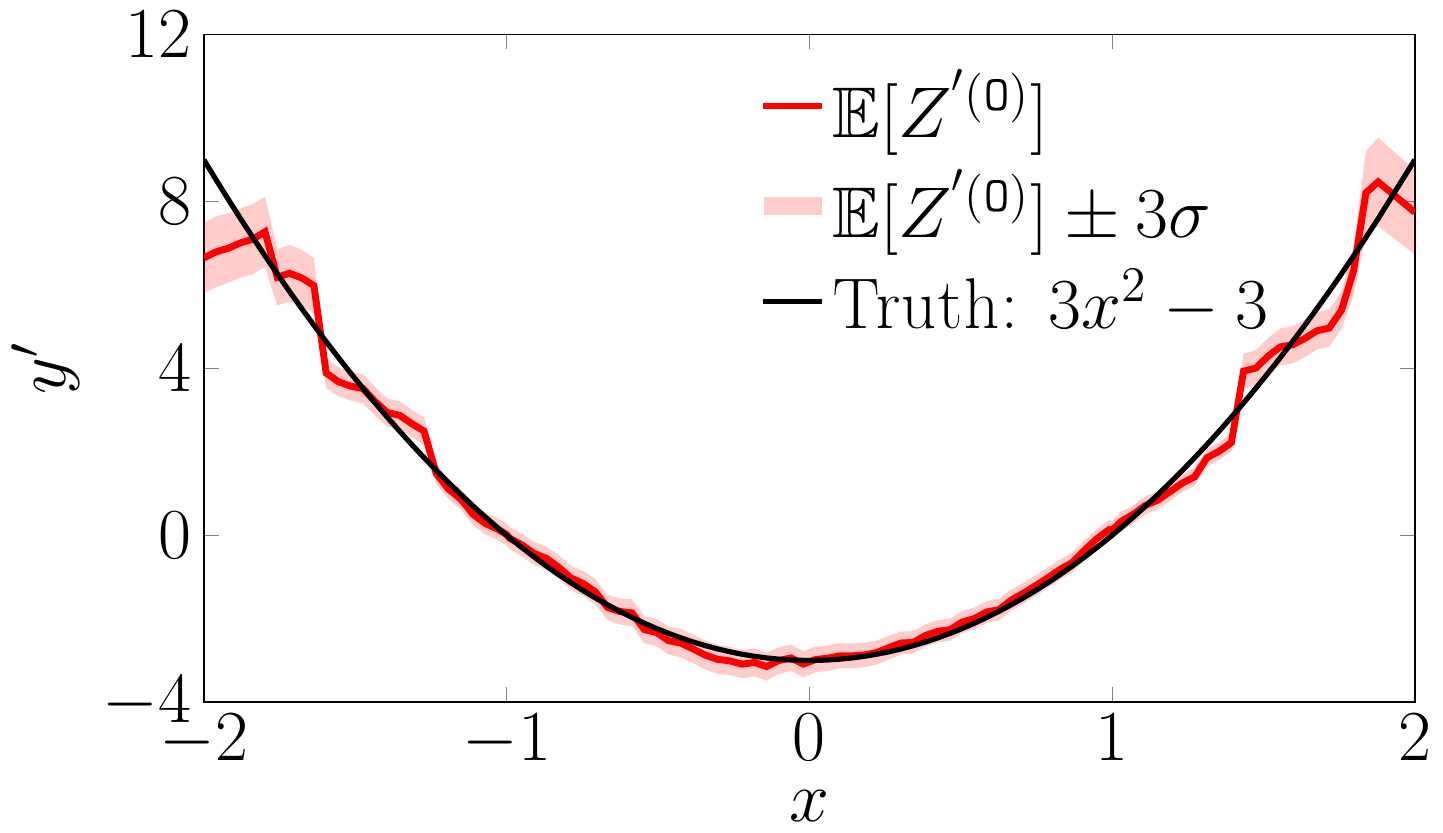}}~~\subfloat[TAGI derivative covariance]{\includegraphics[width=62mm]{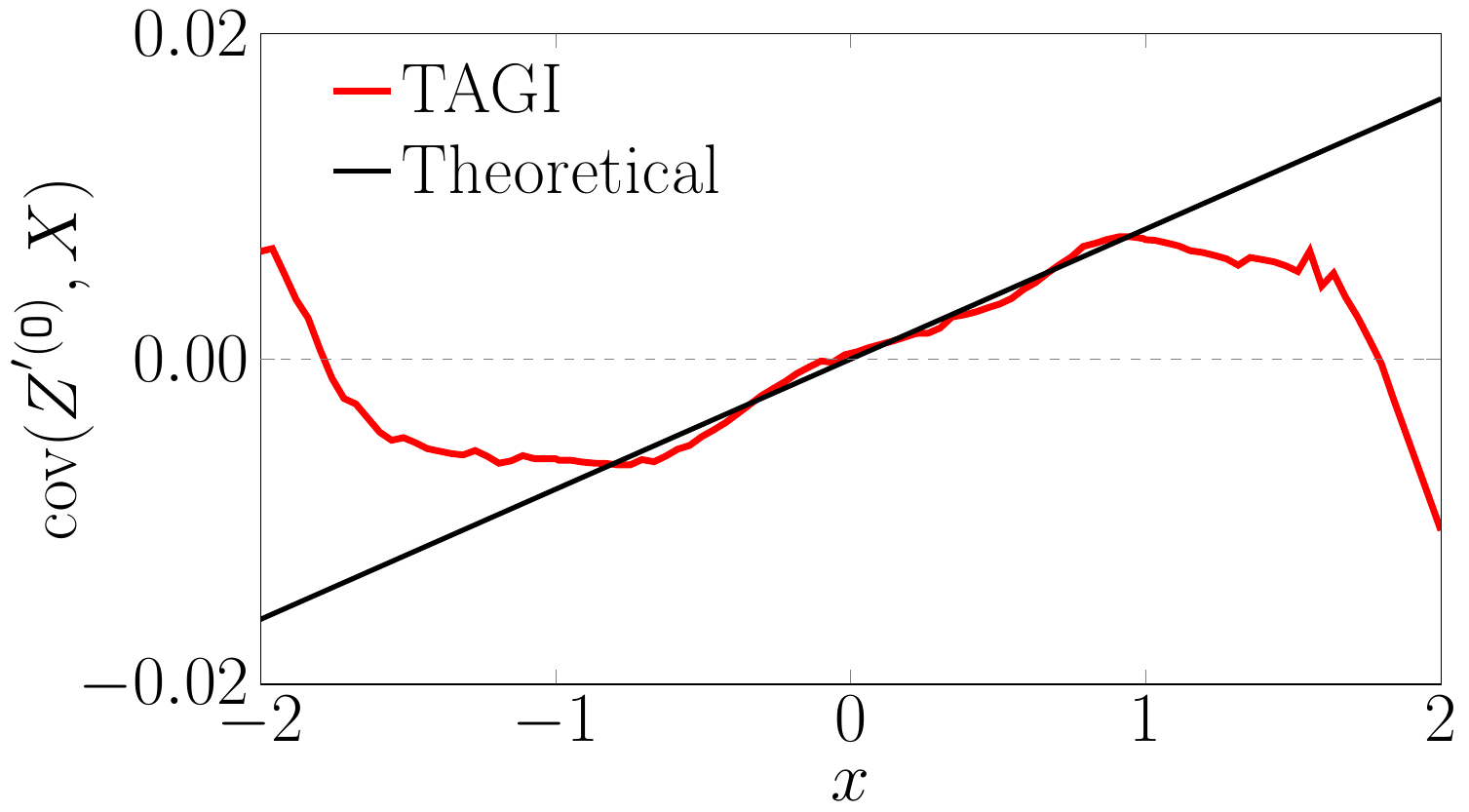}}
\caption{1D optimization problem}
\label{F:1D}
\end{figure}

We illustrate the inference-based optimization scheme on a 1D toy problem for $y=x^3-3x+v$ as depicted in Figure \ref{F:1D}a, where the observation errors $V\sim\mathcal{N}(0,0.1^2)$.  The function approximation obtained using TAGI is presented in Figure \ref{F:1D}b and its derivative in Figure \ref{F:1D}c. We use this toy problem to illustrate how we use the derivative constraint $\alpha$ in order to reach either the local maximum at $x=-1$ or the local minimum at $x=+1$.

Table \ref{tab_opt} presents the optimal location ${\mu}_{{X}|{z}^{'}}$ found by TAGI depending on the starting location $\mu_{{X}}^{\texttt{0}}$ and whether or not a derivative constraint $\alpha$ is employed. Note that for all cases, the initial input variance is set to $\sigma_X=0.01$. The results show that when no derivative constraint $\alpha$ is employed, the optimal value reached correspond to either a maximum or a minimum, depending on the starting location $\mu_{{X}}^{\texttt{0}}$. More specifically, the inference will lead to the maximum or minimum associated within the region where the sign of the covariance is the same as for the starting location $\mu_{{X}}^{\texttt{0}}$, as depicted in Figure \ref{F:1D}d. A positive derivative constrain $\alpha$ leads to the local minimum whether starting in a region having a positive or negative covariance. On the other hand,  a negative constrain leads to the local maximum. Note that whether or not we use a derivative constrain $\alpha$, TAGI will fail to infer the local maximum at $x=-1$ while starting at a value such as $x=1.9$, because the sign of the covariance estimated using TAGI is incorrect so that the optimal location inferred will be pushed beyond the value $x=+2$. This example illustrates a limitation of TAGI's inference-based optimization scheme where, like for gradient-based approaches, the starting location $\mu_{{X}}^{\texttt{0}}$ matters.
\begin{table}[htbp]
\centering
\caption{Comparison of the optimal values obtained ${\mu}_{{X}|{z}^{'}}$ depending on the starting location $\mu_{{X}}^{\texttt{0}}$ and whether or not a derivative constraint $\alpha$ is employed.}
\scalebox{1}{\small
\begin{tabular}{cccc}
    	\addlinespace
    \toprule
    $\mu_{{X}}^{\texttt{0}}$&$\alpha$&\begin{tabular}{c}Update\\equation\end{tabular}&${\mu}_{{X}|{z}^{'}}$\\[-2pt] 
    \cmidrule{1-4}
$0.25$&N/A&Eq. \ref{eq_fwdtagi}&0.965\\[2pt]
$-0.25$&N/A&Eq. \ref{eq_fwdtagi}&$-0.992$\\[2pt]
\cmidrule{1-4}
$0.25$&$+1$&Eq. \ref{eq_opttagi}&$-0.993$\\[2pt]
$-0.25$&$+1$&Eq. \ref{eq_opttagi}&$-0.992$\\[2pt]
\cmidrule{1-4}
$0.25$&$-1$&Eq. \ref{eq_opttagi}&0.965\\[2pt]
$-0.25$&$-1$&Eq. \ref{eq_opttagi}&0.965\\[2pt]
\bottomrule
\end{tabular}}
\label{tab_opt}
\end{table}

Although this optimization problem is trivial as it involves only one dimension, it showcases how the inference capability of TAGI can be leveraged in order to solve optimization tasks. The next section will build on that capacity in order to tackle continuous-actions reinforcement learning problems which involve optimization in higher-dimensional spaces.

\section{Continuous-Actions RL through Inference}\label{S:TAGI_PN}
This section presents how to perform continuous-actions reinforcement learning (RL) by leveraging hidden-state inference.
For both categorical and continuous actions RL frameworks, an agent's goal is to maximize the expected value conditional on an action $a$. For categorical actions, this can be achieved through the explicit evaluation of the expected value for each action and the selection of the optimal one. In the case of continuous actions, it is not possible nor desirable to evaluate the expected value associated with all possible actions; one thus face a continuous optimization problem. In deep-RL methods such as advantage actor critic (A2C) \cite{mnih2016asynchronous} and proximal policy optimization (PPO) \cite{schulman2017proximal}, this optimization is tackled using gradient ascent approaches. Here, we rely instead on the method presented in \S\ref{S:opt}  to identify the optimal action through inference.
  
For typical RL problems, the environment's state at a time $t$ and $t+1$ are $\{\bm{s}, \bm{s}'\}\in\mathbb{R}^{\mathtt{N}^2}$, and the expected utility conditional on the actions $\bm{a}\in \mathbb{R}^{\mathtt{A}}$ and states $\bm{s}$ is defined by the action-value function $q(\bm{s},\bm{a})\in\mathbb{R}^{1}$. Figure \ref{FIG:CNN_RL_cont}a presents the directed acyclic graph (DAG) describing the interconnectivity in a neural network capable of modelling a policy network, i.e., the dependency between the actions $\bm{a}$ and the states $\bm{s}$.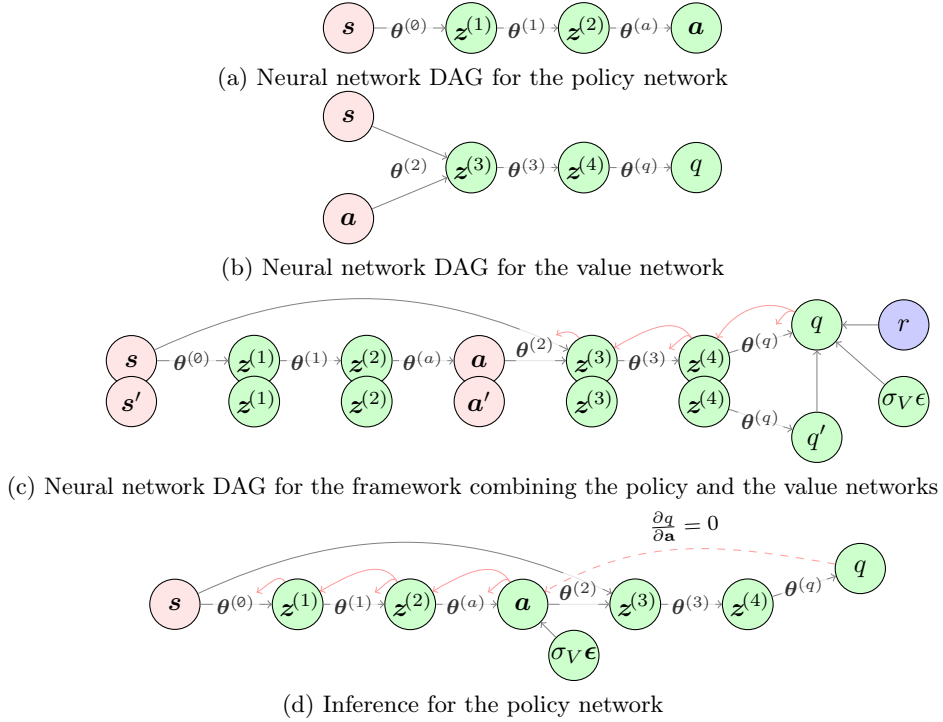
\begin{figure}[h!]
\centering
\subfloat[][Neural network DAG for the policy network]{
\tikzstyle{input}=[draw,fill=red!10,circle,minimum size=19pt,inner sep=0pt]
\tikzstyle{hidden}=[draw,fill=green!20,circle,minimum size=19pt,inner sep=0pt]
\tikzstyle{cnn}=[draw,fill=cyan!40,rectangle,minimum size=19pt,inner sep=3pt]
\tikzstyle{obs}=[draw,fill=blue!20,circle,minimum size=19pt,inner sep=0pt]
\tikzstyle{cte}=[draw=none,circle,minimum size=17pt,inner sep=0pt]
\tikzstyle{error}=[draw,fill=yellow!20,circle,minimum size=19pt,inner sep=0pt]
\tikzstyle{action}=[draw,fill=red!60,circle,minimum size=19pt,inner sep=0pt]
\tikzstyle{inference}=[->,line width=0.25pt,draw=blue!40]
\tikzstyle{label}=[,opacity=0.8,fill=white,circle,inner sep=0.5pt]
\tikzstyle{stateTransition}=[->,line width=0.25pt,draw=black!50]

\begin{tikzpicture}[scale=1.36]\small
    \node (s)[input]   at (0, 0) {$\bm{s}$};
    \node (z1)[hidden] at (1.2, 0) {$\,\bm{z}^{\!(1)}$};
    \node (z2)[hidden] at (2.3, 0) {$\,\bm{z}^{\!(2)}$};
    \node (a)[hidden] at (3.4, 0) {$\bm{a}$};

     \draw[stateTransition] (s) -- (z1) node [label,midway] {\scriptsize$\bm{\theta}^{(\texttt{0})}$};
    \draw[stateTransition] (z1) -- (z2) node [label,midway] {\scriptsize$\bm{\theta}^{(1)}$};

   \draw[stateTransition] (z2) --  (a) node [label,midway] {\scriptsize$\bm{\theta}^{(a)}$};

\end{tikzpicture}}\\[-2pt]
\subfloat[][Neural network DAG for the value network]{
\tikzstyle{input}=[draw,fill=red!10,circle,minimum size=19pt,inner sep=0pt]
\tikzstyle{hidden}=[draw,fill=green!20,circle,minimum size=19pt,inner sep=0pt]
\tikzstyle{cnn}=[draw,fill=cyan!40,rectangle,minimum size=19pt,inner sep=3pt]
\tikzstyle{obs}=[draw,fill=blue!20,circle,minimum size=19pt,inner sep=0pt]
\tikzstyle{cte}=[draw=none,circle,minimum size=17pt,inner sep=0pt]
\tikzstyle{error}=[draw,fill=yellow!20,circle,minimum size=19pt,inner sep=0pt]
\tikzstyle{action}=[draw,fill=red!60,circle,minimum size=19pt,inner sep=0pt]
\tikzstyle{inference}=[->,line width=0.25pt,draw=blue!40]
\tikzstyle{label}=[,opacity=0.8,fill=white,circle,inner sep=0.5pt]
\tikzstyle{stateTransition}=[->,line width=0.25pt,draw=black!50]

\begin{tikzpicture}[scale=1.36]\small
    \node (s)[input]   at (0, 0.5) {$\bm{s}$};
    \node (a)[input] at (0, -0.5) {$\bm{a}$};
    \node (z1)[hidden] at (1.2, 0) {$\,\bm{z}^{\!(3)}$};
    \node (z2)[hidden] at (2.3, 0) {$\,\bm{z}^{\!(4)}$};
    \node (q)[hidden] at (3.4, 0) {${q}$};

     \draw[stateTransition] (s) -- (z1) node [label,midway,yshift=-3.5mm] {\scriptsize$\bm{\theta}^{(2)}$};
          \draw[stateTransition] (a) -- (z1);
    \draw[stateTransition] (z1) -- (z2) node [label,midway] {\scriptsize$\bm{\theta}^{(3)}$};

   \draw[stateTransition] (z2) --  (q) node [label,midway] {\scriptsize$\bm{\theta}^{(q)}$};

\end{tikzpicture}}\\[-2pt]
\subfloat[][Neural network DAG for the framework combining the policy and the value networks]{\tikzstyle{input}=[draw,fill=red!10,circle,minimum size=19pt,inner sep=0pt]
\tikzstyle{hidden}=[draw,fill=green!20,circle,minimum size=19pt,inner sep=0pt]
\tikzstyle{cnn}=[draw,fill=cyan!40,rectangle,minimum size=19pt,inner sep=3pt]
\tikzstyle{obs}=[draw,fill=blue!20,circle,minimum size=19pt,inner sep=0pt]
\tikzstyle{cte}=[draw=none,circle,minimum size=17pt,inner sep=0pt]
\tikzstyle{error}=[draw,fill=yellow!20,circle,minimum size=19pt,inner sep=0pt]
\tikzstyle{action}=[draw,fill=red!60,circle,minimum size=19pt,inner sep=0pt]
\tikzstyle{inference}=[->,line width=0.25pt,draw=red!40]
\tikzstyle{label}=[,opacity=0.8,fill=white,circle,inner sep=0.5pt]
\tikzstyle{stateTransition}=[->,line width=0.25pt,draw=black!50]

\begin{tikzpicture}[scale=1.36]\small
    \node (s)[input]   at (1.2, 0.2) {$\bm{s}$};
    \node (z1)[hidden] at (2.4, 0.2) {$\,\bm{z}^{\!(1)}$};
    \node (z2)[hidden] at (3.5, 0.2) {$\,\bm{z}^{\!(2)}$};
    \node (sp)[input]   at (1.2, -0.2) {$\bm{s}'$};
    \node (z1p)[hidden] at (2.4, -0.2) {$\,\bm{z}^{\!(1)}$};
    \node (z2p)[hidden] at (3.5, -0.2) {$\,\bm{z}^{\!(2)}$};
    
    \node (q)[hidden] at (7.9,0.55) {${q}$};
    \node (v)[hidden] at (8.75,-0.2) {$\sigma_V \epsilon$};
    \node (qp)[hidden] at (7.9,-0.55) {${q}'$};

    \node (a)[input] at (4.6, 0.2) {$\bm{a}$};
    \node (z3)[hidden] at (5.7, 0.2) {$\,\bm{z}^{\!(3)}$};
    \node (z4)[hidden] at (6.8, 0.2) {$\,\bm{z}^{\!(4)}$};
    \node (ap)[input] at (4.6, -0.2) {$\bm{a}'$};
    \node (z3p)[hidden] at (5.7, -0.2) {$\,\bm{z}^{\!(3)}$};
    \node (z4p)[hidden] at (6.8, -0.2) {$\,\bm{z}^{\!(4)}$};

    \node (r)[obs] at (8.75,0.55) {${r}$};

    \draw[stateTransition] (s) -- (z1) node (t0)[label,midway] {\scriptsize$\bm{\theta}^{(\texttt{0})}$};
    \draw[stateTransition] (z1) -- (z2) node (t1)[label,midway] {\scriptsize$\bm{\theta}^{(1)}$};
     \draw[stateTransition] (s) to  [out=25,in=155] (z3) node (tq1)[label, xshift=-27mm,yshift=7mm] {};
     \draw[stateTransition] (r) --  (q);
     
     
     \draw[stateTransition] (v) -- (q);
     \draw[stateTransition] (qp) -- (q);


\draw[stateTransition] (z2) --  (a) node (ta)[label,midway] {\scriptsize$\bm{\theta}^{(a)}$};
\draw[stateTransition] (a) -- (z3) node (t2)[label, above=-4pt, pos=0.5] {\scriptsize$\bm{\theta}^{(2)}$};
\draw[stateTransition] (z3) -- (z4) node (t3)[label,midway] {\scriptsize$\bm{\theta}^{(3)}$};
\draw[stateTransition] (z4) -- (q) node (tq2)[label,midway] {\scriptsize$\bm{\theta}^{(q)}$};
\draw[stateTransition] (z4p) -- (qp) node [label,midway] {\scriptsize$\bm{\theta}^{(q)}$};

	\draw[inference] (q) to [out=150,in=50]  (tq2);
	\draw[inference] (q) to [out=150,in=60]  (z4);
	\draw[inference] (z4) to [out=120,in=30] (z3);
	\draw[inference] (z4) to [out=120,in=30] (t3);
	\draw[inference] (z3) to [out=120,in=30] (t2);
	


\end{tikzpicture}}\\[-2pt]
\subfloat[][Inference for the policy network]{\tikzstyle{input}=[draw,fill=red!10,circle,minimum size=19pt,inner sep=0pt]
\tikzstyle{hidden}=[draw,fill=green!20,circle,minimum size=19pt,inner sep=0pt]
\tikzstyle{cnn}=[draw,fill=cyan!40,rectangle,minimum size=19pt,inner sep=3pt]
\tikzstyle{obs}=[draw,fill=blue!20,circle,minimum size=19pt,inner sep=0pt]
\tikzstyle{cte}=[draw=none,circle,minimum size=17pt,inner sep=0pt]
\tikzstyle{error}=[draw,fill=yellow!20,circle,minimum size=19pt,inner sep=0pt]
\tikzstyle{action}=[draw,fill=red!60,circle,minimum size=19pt,inner sep=0pt]
\tikzstyle{inference}=[->,line width=0.25pt,draw=red!40]
\tikzstyle{label}=[,opacity=0.8,fill=white,circle,inner sep=0.5pt]
\tikzstyle{stateTransition}=[->,line width=0.25pt,draw=black!50]

\begin{tikzpicture}[scale=1.36]\small
    \node (s)[input]   at (1.2, 0.2) {$\bm{s}$};
    \node (z1)[hidden] at (2.4, 0.2) {$\,\bm{z}^{\!(1)}$};
    \node (z2)[hidden] at (3.5, 0.2) {$\,\bm{z}^{\!(2)}$};    
    \node (q)[hidden] at (7.9,0.55) {${q}$};

    \node (a)[hidden] at (4.6, 0.2) {$\bm{a}$};
    \node (z3)[hidden] at (5.7, 0.2) {$\,\bm{z}^{\!(3)}$};
    \node (z4)[hidden] at (6.8, 0.2) {$\,\bm{z}^{\!(4)}$};

    \draw[stateTransition] (s) -- (z1) node (t0)[label,midway] {\scriptsize$\bm{\theta}^{(\texttt{0})}$};
    \draw[stateTransition] (z1) -- (z2) node (t1)[label,midway] {\scriptsize$\bm{\theta}^{(1)}$};
     \draw[stateTransition] (s) to  [out=25,in=155] (z3) node (tq1)[label, xshift=-27mm,yshift=7mm] {};
\draw[stateTransition] (z2) --  (a) node (ta)[label,midway] {\scriptsize$\bm{\theta}^{(a)}$};
\draw[stateTransition] (a) -- (z3) node (t2)[label, above=-4pt, pos=0.5] {\scriptsize$\bm{\theta}^{(2)}$};
\draw[stateTransition] (z3) -- (z4) node (t3)[label,midway] {\scriptsize$\bm{\theta}^{(3)}$};
\draw[stateTransition] (z4) -- (q) node (tq2)[label,midway] {\scriptsize$\bm{\theta}^{(q)}$};
	\draw[inference] (a) to [out=120,in=30] (z2);
	\draw[inference] (a) to [out=120,in=30] (ta);
	\draw[inference] (z2) to [out=120,in=30] (z1);
	\draw[inference] (z2) to [out=120,in=30] (t1);
	\draw[inference] (z1) to [out=120,in=30] (t0);
	\draw[inference,dashed] (q)  to [out=170,in=30] node[sloped, above]{\scriptsize$\tfrac{\partial q}{\partial \mathbf{a}}=0$}(a);
	
 \node (v)[hidden] at (5.1,-0.3) {$\sigma_V \bm{\epsilon}$};
  \draw[stateTransition] (v) -- (a);
\end{tikzpicture}}
\caption{Graphical representation of a neural network structure for temporal-difference Q-policy learning for continuous actions.}
\label{FIG:CNN_RL_cont}
\end{figure} Figure \ref{FIG:CNN_RL_cont}b presents a similar graph for a value network modelling the dependency between the action-value function $q$, and the actions $\bm{a}$ and states $\bm{s}$. Figure \ref{FIG:CNN_RL_cont}c presents the combination of the value and policy networks from (a) and (b) in a single network that is analogous to the temporal-difference learning framework by Nguyen and Goulet \cite{ha2021analytically}, where  $\{\bm{s},\bm{a}\}$ are the states and action at a time $t$ and $\{\bm{s}',\bm{a}'\}$ the states and action at a time $t+1$. In this graph, the nodes that have been doubled represent the states $\bm{s}$ and $\bm{s}'$ which are both evaluated in a network sharing the same parameters in order to learn from the observation equation
\begin{equation}
q(\bm{s},\bm{a})=r(\bm{s})+\gamma q(\bm{s}',\bm{a'})+\sigma_{V}\epsilon,
\label{EQ:EUTD}
\end{equation}
where $\epsilon$ is a realization from a standard-normal random variable, $r(\bm{s})$ is the reward function, and $\gamma$ is the discount factor.

One particularity in the graph from Figure \ref{FIG:CNN_RL_cont}c is that the actions $\{\bm{a},\bm{a}'\}$ are deterministic inputs (red nodes), as the specific actions at a time $t$ are sampled from their current posterior predictive distribution. The red arrows outline the flow of information during the inference procedure for the components belonging to the value network. Note that the policy network cannot be updated directly because the flow of information  in Figure \ref{FIG:CNN_RL_cont}c is broken by the knowledge of the actions. The component belonging to the policy network are thus updated separately as depicted in Figure \ref{FIG:CNN_RL_cont}d, where the prior for  the actions $\bm{a}$ is computed from the policy network so that 
\begin{equation}
\label{eq:obsPolicy}
\bm{a}=\bm{w}^{(a)}\bm{z}^{(2)}+\bm{b}^{(a)}+\sigma_V \bm{\epsilon},
\end{equation}
and where the inference for the actions uses the constrain on the derivative
\begin{equation}
\bm{a}:\frac{\partial q}{\partial \bm{a}}=0.
\end{equation} 
Algorithm \ref{alg_carl} details an example of implementation for the on-policy reinforcement learning in the context of TAGI.
\begin{figure}[htbp]
\begin{center}
\SetCustomAlgoRuledWidth{12.5cm}
\IncMargin{2pt}
\begin{algorithm}[H]
Define policy network $\mathcal{P}\left(\mathbf{s};\bm\theta^{\mathcal{P}}\right)$, value network $\mathcal{Q}\left(\mathbf{s}, \mathbf{a};\bm\theta^{\mathcal{Q}}\right)$;\\[2pt]
Initialize $\bm\theta^{\mathcal{P}}$, $\bm\theta^{\mathcal{Q}}$, $\sigma_{V}$, horizon $\mathtt{H}$, memory $\mathcal{R}$ to capacity $\mathtt{H}$\\[2pt]
steps = 0;\\[2pt]
 \For{episode $= 1 : \mathtt{E}$}{
Reset environment $\mathbf{s}_{1}$;\\[2pt]
\For{$t= 1 : \mathtt{T}$}{
  steps  = steps  + 1;\\[2pt]
  $\left\{\bm\mu^{\mathbf{A}}_{t}, \bm\Sigma^{\mathbf{A}}_{t}\right\} = \mathcal{P}\left(\mathbf{s}_{t}; \bm\theta^{\mathcal{P}}\right);$\\[2pt]
$\mathbf{a}_{t}: \mathbf{A}_{t}\sim\mathcal{N}(\bm\mu^{\mathbf{A}}_{t}, \bm\Sigma^{\mathbf{A}}_{t})$;\\[2pt]
 $\mathbf{s}_{t+1}, r_{t} = \text{enviroment}(\mathbf{a}_{t});$\\[2pt]
 Store $\{\mathbf{s}_{t}, \mathbf{a}_{t}, r_{t}\}$ in $\mathcal{R}$;\\[2pt]
 \If{$steps \,\, \text{mod}\,\, \mathtt{H}==0$}{
   $\left\{\bm\mu^{\mathbf{A}}_{t+1}, \bm\Sigma^{\mathbf{A}}_{t+1}\right\} = \mathcal{P}\left(\mathbf{s}_{t+1}; \bm\theta^{\mathcal{P}}\right);$\\[2pt]
$\mathbf{a}_{t+1}: \mathbf{A}_{t+1}\sim\mathcal{N}(\bm\mu^{\mathbf{A}}_{t+1}, \bm\Sigma^{\mathbf{A}}_{t+1})$;\\[2pt]
$\left\{\mu^{Q}_{t+1}, (\sigma^{Q}_{t+1})^2\right\} = \mathcal{Q}\left(\mathbf{s}_{t+1}, \mathbf{a}_{t+1};\bm\theta^{\mathcal{Q}}\right)$;\\[4pt]
Take $\mathtt{H}$ samples of $\{\mathbf{s}, \mathbf{a}, r\}$ from $\mathcal{R}$;\\[4pt]
$\mu^{Y}_{\mathtt{H}} = \mu^{Q}_{t+1}; \sigma^{Y}_{\mathtt{H}} =  \sigma^{Q}_{t+1}$;\\[4pt]
\For{$j= \mathtt{H}-1:1$}{
   $\mu^{Y}_{j} = r_{j} + \gamma \mu^{Y}_{j+1} ; (\sigma^{Y}_{j})^2 = \gamma^{2}(\sigma^{Y}_{j+1})^2 + \sigma_{V}^2$;\\[4pt]
  }
Update $\bm\theta^{\mathcal{Q}}$ using TAGI;\\[2pt]
Update $\bm\theta^{\mathcal{P}}$ using TAGI and Algorithm \ref{alg_opt} with the constraint $\tfrac{\partial q}{\partial{\mathbf{a}}}=0$;\\[2pt]
Initialize memory $\mathcal{R}$ to capacity $\mathtt{H}$;\\[2pt]
  }
 }
 }
\caption{Continuous-action reinforcement learning with TAGI}
\label{alg_carl}
\end{algorithm}
\end{center}
\end{figure}

We compare the performance of this on-policy TD reinforcement learning framework for continuous actions with the PPO method \cite{schulman2017proximal}. We perform this comparison on the half-cheetah and inverted pendulum problems from the Mujoco environment \cite{todorov2012mujoco} implemented in OpenAI Gym \cite{brockman2016openai}. For the TAGI-based approach, the Q-value network uses a FNN with three hidden layers of $128$ units. The policy network employs a FNN with two hidden layers of $128$ units.  The standard deviation $\sigma_{V}$ in Equation \ref{EQ:EUTD} and \ref{eq:obsPolicy}  is initialized at $2$ and is decayed each $1024$ steps with a decaying factor of $0.9999$. The minimal standard deviation is $\sigma^{\min}_{V} = 0.3$. These hyperparameters are kept constant for both environments. For the PPO approach, we use the same model architecture for both the policy and value networks as well as the hyper-parameter values from Open AI baselines \cite{baselines}.  During training, the TAGI-based approach uses a single epoch while PPO employs ten. The details for the model architecture and hyper-parameter values are provided in Appendix \ref{SS:A:CRL}. Figure \ref{fig_avgReward} shows the average reward over 100 episodes with respect to the number of steps for both environments.
\begin{figure}[htbp]
\begin{center}
\subfloat[HalfCheetah-v2]{\includegraphics[width=62mm]{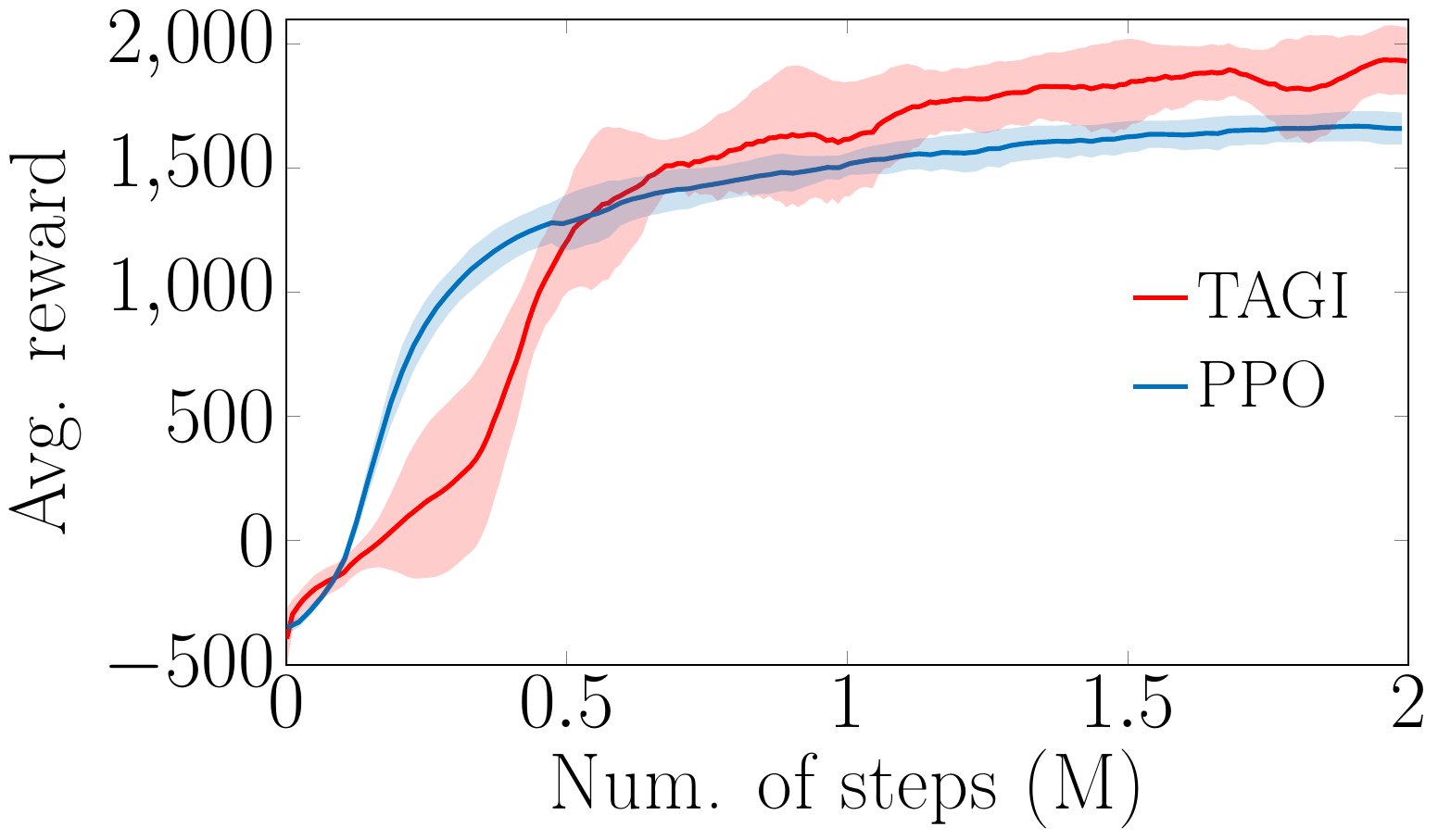}}~~\subfloat[InvertedPendulum-v2]{\includegraphics[width=62mm]{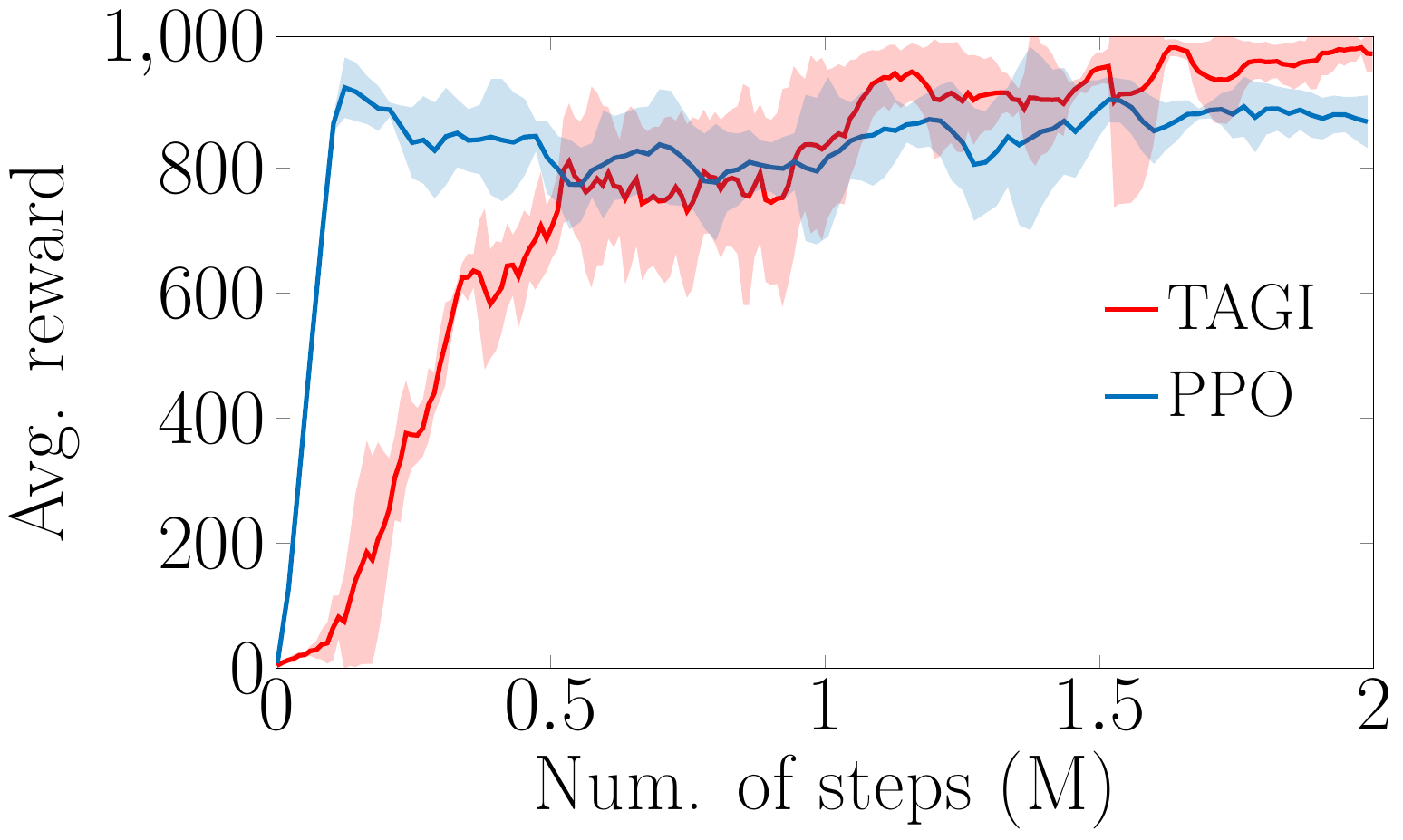}}
\caption{Illustration of average reward over 100 episodes of five random runs for two millions time steps.}
\label{fig_avgReward}
\end{center}
\end{figure}
Table \ref{tab_avgReward} presents the average reward over the last 100 episodes for both environments. 
\begin{table}[htbp]
\begin{center}
\caption{Average reward over the last 100 episodes of five random runs for the half-cheetah and inverted pendulum experiments. TAGI: Tractable Approximate Gaussian Inference; PPO: Proximal Policy Optimization.}
\label{tab_avgReward}
\label{t:offpolicy_results}
\scalebox{1}{ \!\!\!\setlength{\tabcolsep}{3.5pt}
\begin{tabular}{lcc}    	\addlinespace
    \toprule
Method&Half-cheetah& Inverted pendulum\\[2pt]
\cmidrule{1-3}
\multirow{1}{*}{TAGI}&1934 $\pm$ 131&983 $\pm$ 30\\[2pt]
PPO& 1649 $\pm$ 48& 887 $\pm$ 42\\[2pt]
\bottomrule
\end{tabular}}
\end{center}
\end{table}
Although PPO initially learns faster, the final results show that TAGI outperforms PPO on both experiments. In addition,  TAGI requires fewer hyper-parameters than PPO (see Table \ref{tab_hprl} in Appendix \ref{SS:A:CRL}). Note that the goal of this experiment is to demonstrate how can inference be leveraged for solving existing problems with a novel approach. The application of TAGI to RL problems is in its early days and it is foreseeable that if more time is invested in exploring new architectures and network configurations, the framework could further exceed the current performance.
 
\section{Conclusion} \label{S:Conclusion}
TAGI provides a novel capacity to perform inference in neural networks. Its applications to adversarial attacks, optimization, and continuous-action reinforcement learning showcase how these tasks, which previously relied on gradient-based optimization methods, can now be approached with analytically tractable inference. The applications presented in this paper are only a subset from the variety of problems that can take advantage of inference, either through the adaptation of existing architectures or through the development of new ones. 

\section*{Acknowledgements}
The first author was financially supported by research grants from Hydro-Quebec, and the Natural Sciences and Engineering Research Council of Canada (NSERC). We would like to thank Magali Goulet and Prof. Mélina Mailhot for having reviewed the equations employed for computing the derivatives. 
\bibliographystyle{abbrv}
\bibliography{Goulet_reference_librairy}

\begin{thebibliography}{10}

\bibitem{akhtar2018threat}
N.~Akhtar and A.~Mian.
\newblock Threat of adversarial attacks on deep learning in computer vision: A
  survey.
\newblock {\em IEEE Access}, 6:14410--14430, 2018.

\bibitem{ardizzone2018analyzing}
L.~Ardizzone, J.~Kruse, C.~Rother, and U.~K{\"o}the.
\newblock Analyzing inverse problems with invertible neural networks.
\newblock In {\em International Conference on Learning Representations}, 2019.

\bibitem{brockman2016openai}
G.~Brockman, V.~Cheung, L.~Pettersson, J.~Schneider, J.~Schulman, J.~Tang, and
  W.~Zaremba.
\newblock Openai gym.
\newblock {\em arXiv preprint arXiv:1606.01540}, 2016.

\bibitem{chen2016infogan}
X.~Chen, Y.~Duan, R.~Houthooft, J.~Schulman, I.~Sutskever, and P.~Abbeel.
\newblock Infogan: Interpretable representation learning by information
  maximizing generative adversarial nets.
\newblock {\em arXiv:1606.03657}, 2016.

\bibitem{baselines}
P.~Dhariwal, C.~Hesse, O.~Klimov, A.~Nichol, M.~Plappert, A.~Radford,
  J.~Schulman, S.~Sidor, Y.~Wu, and P.~Zhokhov.
\newblock Openai baselines.
\newblock \url{https://github.com/openai/baselines}, 2017.

\bibitem{goodfellow2016deep}
I.~Goodfellow, Y.~Bengio, and A.~Courville.
\newblock {\em Deep learning}.
\newblock MIT Press, 2016.

\bibitem{goodfellow2014generative}
I.~Goodfellow, J.~Pouget-Abadie, M.~Mirza, B.~Xu, D.~Warde-Farley, S.~Ozair,
  A.~Courville, and Y.~Bengio.
\newblock Generative adversarial nets.
\newblock {\em Advances in neural information processing systems},
  27:2672--2680, 2014.

\bibitem{43405}
I.~Goodfellow, J.~Shlens, and C.~Szegedy.
\newblock Explaining and harnessing adversarial examples.
\newblock In {\em International Conference on Learning Representations}, 2015.

\bibitem{2020_TAGI_arxiv}
J.-A. Goulet, L.~Nguyen, and S.~Amiri.
\newblock Tractable approximate {Gaussian} inference for {Bayesian} neural
  networks.
\newblock {\em arXiv preprint}, 2020.

\bibitem{goulet2020tractable}
J.-A. Goulet, L.~H. Nguyen, and S.~Amiri.
\newblock Tractable approximate gaussian inference for {B}ayesian neural
  networks.
\newblock {\em arXiv}, (2004.09281, cs.LG), 2020.

\bibitem{kingma2013auto}
D.~P. Kingma and M.~Welling.
\newblock Auto-encoding variational {B}ayes.
\newblock {\em arXiv preprint arXiv:1312.6114}, 2013.

\bibitem{kruse2021benchmarking}
J.~Kruse, L.~Ardizzone, C.~Rother, and U.~K{\"o}the.
\newblock Benchmarking invertible architectures on inverse problems, 2021.

\bibitem{mnih2016asynchronous}
V.~Mnih, A.~P. Badia, M.~Mirza, A.~Graves, T.~Lillicrap, T.~Harley, D.~Silver,
  and K.~Kavukcuoglu.
\newblock Asynchronous methods for deep reinforcement learning.
\newblock In {\em International conference on machine learning}, pages
  1928--1937. PMLR, 2016.

\bibitem{ha2021analytically}
L.~H. Nguyen and J.-A. Goulet.
\newblock Analytically tractable {B}ayesian deep {Q}-learning.
\newblock {\em arXiv preprint arXiv:2106.11086}, 2021.

\bibitem{nguyen2021analytically}
L.~H. Nguyen and J.-A. Goulet.
\newblock Analytically tractable inference in deep neural networks.
\newblock {\em arXiv preprint arXiv:2103.05461}, 2021.

\bibitem{rumelhart1988learning}
D.~E. Rumelhart, G.~E. Hinton, and R.~J. Williams.
\newblock Learning representations by back-propagating errors.
\newblock {\em Nature}, 323:533---536, 1986.

\bibitem{schulman2017proximal}
J.~Schulman, F.~Wolski, P.~Dhariwal, A.~Radford, and O.~Klimov.
\newblock Proximal policy optimization algorithms.
\newblock {\em arXiv preprint arXiv:1707.06347}, 2017.

\bibitem{sutton2011reinforcement}
R.~S. Sutton and A.~G. Barto.
\newblock {\em Reinforcement learning: An introduction}.
\newblock MIT Press, 2nd edition, 2018.

\bibitem{todorov2012mujoco}
E.~Todorov, T.~Erez, and Y.~Tassa.
\newblock Mujoco: A physics engine for model-based control.
\newblock In {\em 2012 IEEE/RSJ International Conference on Intelligent Robots
  and Systems}, pages 5026--5033. IEEE, 2012.

\end{thebibliography}
%
%
%
\newpage
\appendix
\section{Model Architecture and Hyper-parameters}\label{A:AAS}
This appendix contains the specifications for each model architecture in the experiment section. $D$ refers to a layer depth; $W$ refers to a layer width; $H$ refers to the layer height in case of convolutional or pooling layers; $K$ refers to the kernel size; $P$ refers to the convolutional kernel padding; $S$ refers to the convolution stride; $\phi$ refers to the activation function type; ReLU refers to rectified linear unit;
\subsection{Adversarial Attack}
\subsubsection{MNIST}\label{C:mnist10)}
\begin{table}[htbp]
\begin{center}
\caption{Configuration details for the CNN applied to the MNIST adversarial attack.}
\scalebox{1}{ \!\!\!\setlength{\tabcolsep}{3.5pt}
\begin{tabular}{lllllc}    	\addlinespace
    \toprule
Layer&$D\times W \times H$& $K\times K$ &$P$&$S$&$\phi$\\[0pt]   
\cmidrule{1-6}
\multirow{1}{*}{Input}&$1\times28\times28$&-&-&-&-\\[2pt]
Convolutional&$32\times27\times27$ &$4\times4$&$1$&$1$&ReLU\\[2pt]
Pooling&$32\times13\times13$ &$3\times3$&$0$&$2$&-\\[2pt]
Convolutional&$64\times9\times9$ &$5\times5$&$0$&$1$&ReLU\\[2pt]
Pooling&$64\times4\times4$ &$3\times3$&$0$&$2$&-\\[2pt]
Fully connected&$150\times1\times1$ &-&-&-&ReLU\\[2pt]
Output&$11\times1\times1$ &-&-&-&-\\[2pt]
\bottomrule
\end{tabular}}
\end{center}
\end{table}

\newpage
\subsubsection{Cifar10}\label{C:cifar10)}
\begin{table}[h!]
\begin{center}
\caption{Configuration details for the CNN applied to the Cifar10  adversarial attack}
\scalebox{1}{ \!\!\!\setlength{\tabcolsep}{3.5pt}
\begin{tabular}{lllllc}    	\addlinespace
    \toprule
Layer&$D\times W \times H$& $K\times K$ &$P$&$S$&$\phi$\\[0pt]   
\cmidrule{1-6}
\multirow{1}{*}{Input}&$3\times32\times32$&-&-&-&-\\[2pt]
Convolutional&$32\times32\times32$ &$5\times5$&$2$&$1$&ReLU\\[2pt]
Pooling&$32\times16\times16$ &$3\times3$&$1$&$2$&-\\[2pt]
Convolutional&$32\times16\times16$ &$5\times5$&$2$&$1$&ReLU\\[2pt]
Average pooling&$32\times8\times8$ &$3\times3$&$1$&$2$&-\\[2pt]
Convolutional&$64\times8\times8$ &$5\times5$&$2$&$1$&ReLU\\[2pt]
Average pooling&$64\times4\times4$ &$3\times3$&$1$&$2$&-\\[2pt]
Fully connected&$64\times1\times1$ &-&-&-&ReLU\\[2pt]
Output&$11\times1\times1$ &-&-&-&-\\[2pt]
\bottomrule
\end{tabular}}
\end{center}
\end{table}

\subsection{Optimization}
\begin{table}[htbp]
\begin{center}
\caption{Configuration details for the feedforward neural network applied to 1D example.}
\scalebox{1}{ \!\!\!\setlength{\tabcolsep}{3.5pt}
\begin{tabular}{lllllc}    	\addlinespace
    \toprule
Layer&$D\times W \times H$& $K\times K$ &$P$&$S$&$\phi$\\[0pt]   
\cmidrule{1-6}
\multirow{1}{*}{Input}&$1\times1\times1$&-&-&-&-\\[2pt]
Fully connected&$64\times1\times1$ &-&-&-&Tanh\\[2pt]
Fully connected&$64\times1\times1$ &-&-&-&ReLU\\[2pt]
Output&$1\times1\times1$ &-&-&-&-\\[2pt]
\bottomrule
\end{tabular}}
\end{center}
\end{table}


\newpage
\subsection{Continuous-Action Reinforcement Learning}\label{SS:A:CRL}
For the half-cheatah environment, the number of states $N_{s}$ is $17$ and the number of actions $N_{a}$ is 6. For the inverted pendulum environment, the number of states $N_{s}$ is $4$ and the number of actions $N_{a}$ is 1. 
\begin{table}[htbp]
\begin{center}
\caption{Configuration details for the policy network. $N_{s}$ is the number of states; $N_{a}$ is the number of actions.}
\scalebox{1}{ \!\!\!\setlength{\tabcolsep}{3.5pt}
\begin{tabular}{lllllc}    	\addlinespace
    \toprule
Layer&$D\times W \times H$& $K\times K$ &$P$&$S$&$\phi$\\[0pt]   
\cmidrule{1-6}
\multirow{1}{*}{Input}&$N_{s}\times1\times1$&-&-&-&-\\[2pt]
Fully connected&$128\times1\times1$ &-&-&-&ReLU\\[2pt]
Fully connected&$128\times1\times1$ &-&-&-&ReLU\\[2pt]
Output&$N_{a}\times1\times1$ &-&-&-&Tanh\\[2pt]
\bottomrule
\end{tabular}}
\end{center}
\end{table}
\begin{table}[htbp]
\begin{center}
\caption{Configuration details for the Q-value network. $N_{s}$ is the number of states.}
\scalebox{1}{ \!\!\!\setlength{\tabcolsep}{3.5pt}
\begin{tabular}{lllllc}    	\addlinespace
    \toprule
Layer&$D\times W \times H$& $K\times K$ &$P$&$S$&$\phi$\\[0pt]   
\cmidrule{1-6}
\multirow{1}{*}{Input}&$N_{s}\times1\times1$&-&-&-&-\\[2pt]
Fully connected&$128\times1\times1$ &-&-&-&Tanh\\[2pt]
Fully connected&$128\times1\times1$ &-&-&-&ReLU\\[2pt]
Fully connected&$128\times1\times1$ &-&-&-&ReLU\\[2pt]
Output&$1\times1\times1$ &-&-&-&-\\[2pt]
\bottomrule
\end{tabular}}
\end{center}
\end{table}

\begin{table}[htbp]
\begin{center}
\caption{Hyper-parameters for half-cheetah and inverted pendulum problems.}
\label{tab_hprl}
\scalebox{0.9}{ \!\!\!\setlength{\tabcolsep}{3.5pt}
\begin{tabular}{llll}    	\addlinespace
    \toprule
Method&\#&Hyperparameter&Value\\[0pt]   
\cmidrule{1-4}
\multirow{9}{*}{TAGI}&1&Horizon&1024\\[2pt]
&2& Initial standard deviation for the value function $(\sigma_V)$& 2\\[2pt]
&3&Decay factor $(\eta)$& 0.9999\\[2pt]
&4&Minimal standard deviation for the value function $(\sigma_V^{\min})$& 0.3\\[2pt]
&5&Batch size & 16\\[2pt]
&6&Number of epochs & 1\\[2pt]
&7&Discount $(\gamma)$ & 0.99\\[2pt]
\cmidrule{1-4}
\multirow{18}{*}{PPO}&1&Horizon&2048\\[2pt]
&2&Adam stepsize & $3\times 10^{-4} \times \alpha$\\[2pt]
&3&Adam epsilon & $10^{-5}$\\[2pt]
&4&Adam $\beta_1$ & $0.9$\\[2pt]
&5&Adam $\beta_2$ & $0.999$\\[2pt]
&6&Batch size & 32\\[2pt]
&7&Number of epochs & 10\\[2pt]
&8&Discount $(\gamma)$ & 0.99\\[2pt]
&9& Generalized advantage estimation parameter $(\lambda)$&0.95\\[2pt]
&10&Clipping parameter $(\epsilon)$&$0.2\times \alpha$\\[2pt]
&11&Value function loss coefficient $(c_{1})$ & 1\\[2pt]
&12&Entropy coefficient $(c_{2})$ & 0.0\\[2pt]
&13&Gradient norm clipping coefficient & 0.5\\[2pt]
&14&$\alpha$&LinearAnneal$(1, 0)$\\[2pt]
\bottomrule
\end{tabular}}
\end{center}
\end{table}

\newpage
\section{Partial Derivative in TAGI Neural Networks}\label{A:DTAGI}
\subsection{TAGI Neural Networks}
In a feedforward neural network, the hidden state at a given layer $l+1$ is defined as
\begin{equation}
\label{eq_nnhs}
Z_{i}^{(l+1)} = \displaystyle\sum_{k}W_{ik}^{(l)}\phi(Z_{k}^{(l)}) + B_{i}^{(l)}, ~\forall i\in[1, \mathtt{A}^{(l+1)}],\, \forall k\in[1, \mathtt{A}^{(l)}],\forall l\in[1, \mathtt{L}]
\end{equation}
where $\phi(.)$ is the activation function, $\{W, B\}$ are the unkown parameters of the neural network, i.e. weight and bias, $\mathtt{A}^{(l)}$, is the number of hidden units in layer $l$ and $\mathtt{L}$ is the number of hidden layers. We define the activation unit $A=\phi(Z)$. In the context of TAGI, $Z, W$, and $B$ are assumed to be Gaussian random variables and 
\begin{equation}
\label{eq_idpApt}
\begin{array}{rcl}
W_{ik}^{(l)}&\independent &W_{np}^{(m)}\independent B_{i}^{(l)}\independent B_{n}^{(m)}, ~\forall m \in[1, \mathtt{L}], \forall n\in[1, \mathtt{A}^{(m+1)}], \forall p\in[1, \mathtt{A}^{(m)}]\\[8pt]
Z_{t}^{(l-1)}&\independent&Z_{i}^{(l+1)}, ~\forall t\in[1, \mathtt{A}^{(l-1)}]\\[8pt]
Z_{k}^{(l)}&\independent& Z_{q}^{(l)},~\forall q\in[1, \mathtt{A}^{(l)}] \,\,\text{and}\,\, k\neq q\\[8pt]
Z_{k}^{(l)}&\independent& W_{ik}^{(l)} \independent B_{i}^{(l)}.
\end{array}
\end{equation}
In addition, we apply the locally linearized activation function $\tilde{\phi}(.)$ to the hidden state in order to obtain the probability density function for the output of $\phi(.)$ so that 
\begin{equation}
\label{eq_lra}
\phi(Z_{k}^{(l)}) = J_{k}^{(l)}\left(Z_{k}^{(l)}-\mathbb{E}\left[Z_{k}^{(l)}\right]\right) + \phi\left(\mathbb{E}\left[Z_{k}^{(l)}\right]\right),
\end{equation}
where $J_{k}^{(l)} = \nabla_{z}\phi\left(\mathbb{E}\left[Z_{k}^{(l)}\right]\right)$. 

\subsection{Gaussian Multiplication Approximation (GMA)}
Assuming $\bm{X}= [X_{1}\,\hdots\,X_{4}]^{\intercal}$ are Gaussian random variables, the GMA formulation had been defined by Goulet, Nguyen and Amiri \cite{goulet2020tractable} as
\begin{eqnarray}
\mathbb{E}[X_1X_2]\!\!\!\!\!&=&\!\!\!\!\!\mu_1\mu_2+\text{cov}(X_1,X_2), \label{eq_m}\\[4pt]
\text{cov}(X_3,X_1X_2)\!\!\!\!\!&=&\!\!\!\!\!\text{cov}(X_1,\!X_3)\mu_2\!+\!\text{cov}(X_2,\!X_3)\mu_1 \label{gma_cov3},\\[4pt]
\text{cov}(X_1X_2,X_3X_4)\!\!\!\!\!&=&\!\!\!\!\!\text{cov}(X_1,X_3)\text{cov}(X_2,X_4) \label{gma_cov4} \\[2pt]
\!\!\!\!\!&&\!\!\!\!\!\!\!\!\!\!\!\!\!\!\!\!\!\! +\text{cov}(X_1,X_4)\text{cov}(X_2,X_3)\nonumber  \\[2pt]
\!\!\!\!\!&&\!\!\!\!\!\!\!\!\!\!\!\!\!\!\!\!\!\!+\text{cov}(X_1,X_3)\mu_2\mu_4+\text{cov}(X_1,X_4)\mu_2\mu_3\nonumber  \\
\!\!\!\!\!&&\!\!\!\!\!\!\!\!\!\!\!\!\!\!\!\!\!\!+\text{cov}(X_2,X_3)\mu_1\mu_4+\text{cov}(X_2,X_4)\mu_1\mu_3, \nonumber\\[4pt]
\text{var}(X_1X_2)\!\!\!&=&\!\!\! \sigma_{1}^{2}\sigma_{2}^{2}+\text{cov}(X_1,X_2)^{2}\label{gma_var}\\
\!\!\!\!\!&&\!\!\!\!\!\!\!\!\!\!\!\!\!\!\!\!\!\!+2\text{cov}(X_1,X_2)\mu_{1}\mu_{2}+\sigma_{1}^{2}\mu_{2}^{2}+\sigma_{2}^{2}\mu_{1}^{2}. \nonumber
\end{eqnarray}
\subsection{Partial Derivative Formulations for A Simple Feedforward Neural Network}
\begin{figure}[htbp]
\begin{center}
\scalebox{0.8}{
\tikzstyle{connect}=[-latex, thick,>=latex]
\tikzstyle{shaded}=[draw=black!20,fill=black!20]
\tikzstyle{line} = [draw, -latex,thick,>=latex]

\tikzstyle{circ1}=[circle,minimum width=0.8cm,minimum height=0.8cm, thick, draw =black!100, inner ysep=-2pt,inner xsep=-4pt] 
\tikzstyle{circ2}=[circle, minimum width=0.8cm,minimum height=0.8cm,thick, draw =black!80,fill = green!80!blue!30, fill opacity=0.6,  text opacity=1,inner ysep=-2pt,inner xsep=-4pt] 
\tikzstyle{dummycirc2}=[circle, minimum width=1cm,minimum height=1cm,thick, draw =none] 
\begin{tikzpicture}
\node[circ1](z1) [label=center:$Z_{1}^{(0)}$]{};
\node[circ1](z2) [below=2.8cm of z1,label=center:$Z_{2}^{(0)}$]{};
\node[circ1](z3) [below=2.8cm of z2,label=center:$Z_{3}^{(0)}$]{};
\node[circ1](z4) [below=2.8cm of z3,label=center:$Z_{4}^{(0)}$]{};
\node[circ2](a1)[right = 0.7cm of z1]{$A_{1}^{(0)}$};
\node[circ2](a2)[right = 0.7cm of z2]{$A_{2}^{(0)}$};
\node[circ2](a3)[right = 0.7cm of z3]{$A_{3}^{(0)}$};
\node[circ2](a4)[right = 0.7cm of z4]{$A_{4}^{(0)}$};
\node[circ1](z5) [below right = 1 cm and 3 cm of a1, label=center:$Z_{1}^{(1)}$]{};
\node[circ1](z6) [below = 2.8cm of z5, label=center:$Z_{2}^{(1)}$]{};
\node[circ1](z7) [below = 2.8cm of z6, label=center:$Z_{3}^{(1)}$]{};
\node[circ2](a5)[right = 0.7cm of z5]{$A_{1}^{(1)}$};
\node[circ2](a6)[right = 0.7cm of z6]{$A_{2}^{(1)}$};
\node[circ2](a7)[right = 0.7cm of z7]{$A_{3}^{(1)}$};
\node[circ1](z8) [below right = 0.2 cm and 2 cm of a5, label=center:$Z_{1}^{(2)}$]{};
\node[circ1](z9) [below = 4cm of z8, label=center:$Z_{2}^{(2)}$]{};
\node[circ2](a8)[right = 0.7cm of z8]{$A_{2}^{(2)}$};
\node[circ2](a9)[right = 0.7cm of z9]{$A_{3}^{(2)}$};
\node[circ1](z10) [below right=1.7cm and 1cm of a8,label=center:$Z_{1}^{(3)}$]{};
\node[circ2](a10) [right=0.7cm of z10,label=center:$A_{1}^{(3)}$]{};
\node[dummycirc2](b1) [above left = 1cm and 0.5cm of z5, label=center:$\mathbf{B}^{(0)}$]{};
\node[dummycirc2](b2) [above left = 1cm and 0.3cm of z8, label=center:$\mathbf{B}^{(1)}$]{};
\node[dummycirc2](b3) [above left = 1cm and -0.5cm of z10, label=center:$\mathbf{B}^{(2)}$]{};
\path  (z1) 	edge[connect, very thin, draw=black!50]node[fill=white]{$\phi$} (a1)
	 (z2) 		edge[connect, very thin, draw=black!50]node[fill=white]{$\phi$} (a2)
	 (z3) 		edge[connect, very thin, draw=black!50]node[fill=white]{$\phi$} (a3)
	 (z4) 		edge[connect, very thin, draw=black!50]node[fill=white]{$\phi$} (a4)
	 (a1) 		edge[connect, very thin, draw=black!50]node[ pos=0.18, sloped,fill=white]{$W_{11}^{(0)}$} (z5)
	 (a2) 		edge[connect, very thin, draw=black!50]node[pos=0.2, sloped,fill=white]{$W_{12}^{(0)}$} (z5)
	 (a3) 		edge[connect, very thin, draw=black!50]node[pos=0.17, sloped,fill=white]{$W_{13}^{(0)}$} (z5)
	 (a4) 		edge[connect, very thin, draw=black!50]node[pos=0.18, sloped,fill=white]{$W_{14}^{(0)}$} (z5)
	 
	 (a1) 		edge[connect, very thin, draw=black!50]node[pos=0.21, sloped,fill=white]{$W_{21}^{(0)}$} (z6)
	 (a2) 		edge[connect, very thin, draw=black!50]node[pos=0.25, sloped,fill=white]{$W_{22}^{(0)}$} (z6)
	 (a3)  	edge[connect, very thin, draw=black!50]node[pos=0.17, sloped,fill=white]{$W_{23}^{(0)}$} (z6)
	 (a4) 		edge[connect, very thin, draw=black!50]node[pos=0.1, sloped,fill=white]{$W_{24}^{(0)}$} (z6)
	 
	 (a1) 		edge[connect, very thin, draw=black!50]node[pos=0.22, sloped,fill=white]{$W_{31}^{(0)}$} (z7)
	 (a2) 		edge[connect, very thin, draw=black!50]node[pos=0.25, sloped,fill=white]{$W_{32}^{(0)}$} (z7)
	 (a3)  	edge[connect, very thin, draw=black!50]node[pos=0.21, sloped,fill=white]{$W_{33}^{(0)}$} (z7)
	 (a4) 		edge[connect, very thin, draw=black!50]node[pos=0.45, sloped,fill=white]{$W_{34}^{(0)}$} (z7)
	 
	 (z5) 		edge[connect, very thin, draw=black!50]node[fill=white]{$\phi$} (a5)
	 (z6) 		edge[connect, very thin, draw=black!50]node[fill=white]{$\phi$} (a6)
	 (z7) 		edge[connect, very thin, draw=black!50]node[fill=white]{$\phi$} (a7)
	 (a5) 		edge[connect, very thin, draw=black!50]node[pos=0.28, sloped,fill=white]{$W_{11}^{(1)}$}(z8)
	 (a6) 		edge[connect, very thin, draw=black!50]node[pos=0.18, sloped,fill=white]{$W_{12}^{(1)}$}(z8)
	 (a7) 		edge[connect, very thin, draw=black!50]node[pos=0.18, sloped,fill=white]{$W_{13}^{(1)}$}(z8)
	 	 
	 (a5) 		edge[connect, very thin, draw=black!50]node[pos=0.18, sloped,fill=white]{$W_{21}^{(1)}$}(z9)
	 (a6) 		edge[connect, very thin, draw=black!50]node[pos=0.28, sloped,fill=white]{$W_{22}^{(1)}$}(z9)
	 (a7)  	edge[connect, very thin, draw=black!50]node[pos=0.38, sloped,fill=white]{$W_{23}^{(1)}$}(z9)
	 
	 (z8) 		edge[connect, very thin, draw=black!50]node[fill=white]{$\phi$} (a8)
	 (z9) 		edge[connect, very thin, draw=black!50]node[fill=white]{$\phi$} (a9)
	 (a8)		edge[connect, very thin, draw=black!50]node[pos=0.32, sloped,fill=white]{$W_{11}^{(2)}$}(z10)
	 (a9)		edge[connect, very thin, draw=black!50]node[pos=0.32, sloped,fill=white]{$W_{12}^{(2)}$}(z10)
	 (z10)       edge[connect, very thin, draw=black!50]node[fill=white]{$\phi$} (a10)
	 
	 (b1) 		edge[connect, very thin, draw=black!50, red] (z5)
	 (b1) 		edge[connect, very thin, draw=black!50, red] (z6)
	 (b1) 		edge[connect, very thin, draw=black!50, red] (z7)
	 
	 (b2) 		edge[connect, very thin, draw=black!50, red] (z8)
	 (b2) 		edge[connect, very thin, draw=black!50, red] (z9)	 
	 (b3) 		edge[connect, very thin, draw=black!50, red] (z10)
;
\end{tikzpicture}}
\caption{Ilustration of parameters $\bm\theta=\{\mathbf{W}, \mathbf{B}\}$, hidden states $\mathbf{Z}$, and activation units $\mathbf{A}$ associated with a four-layer feedforward neural network.}
\label{fig_4lnn}
\end{center}
\end{figure}
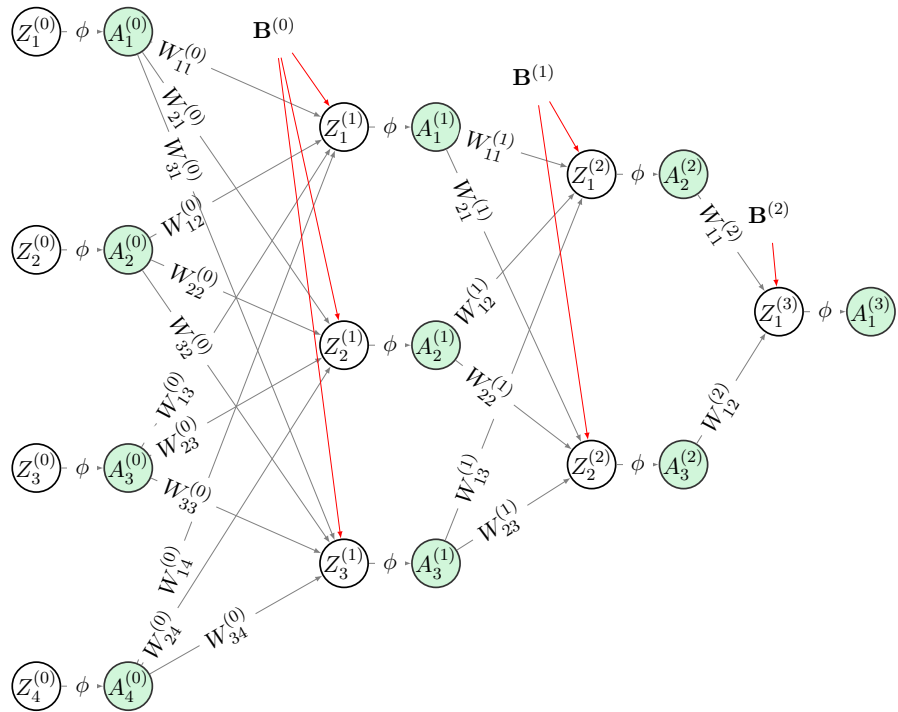

\begin{figure}[htbp]
\centering
\scalebox{0.6}{\rotatebox{-90}{\input{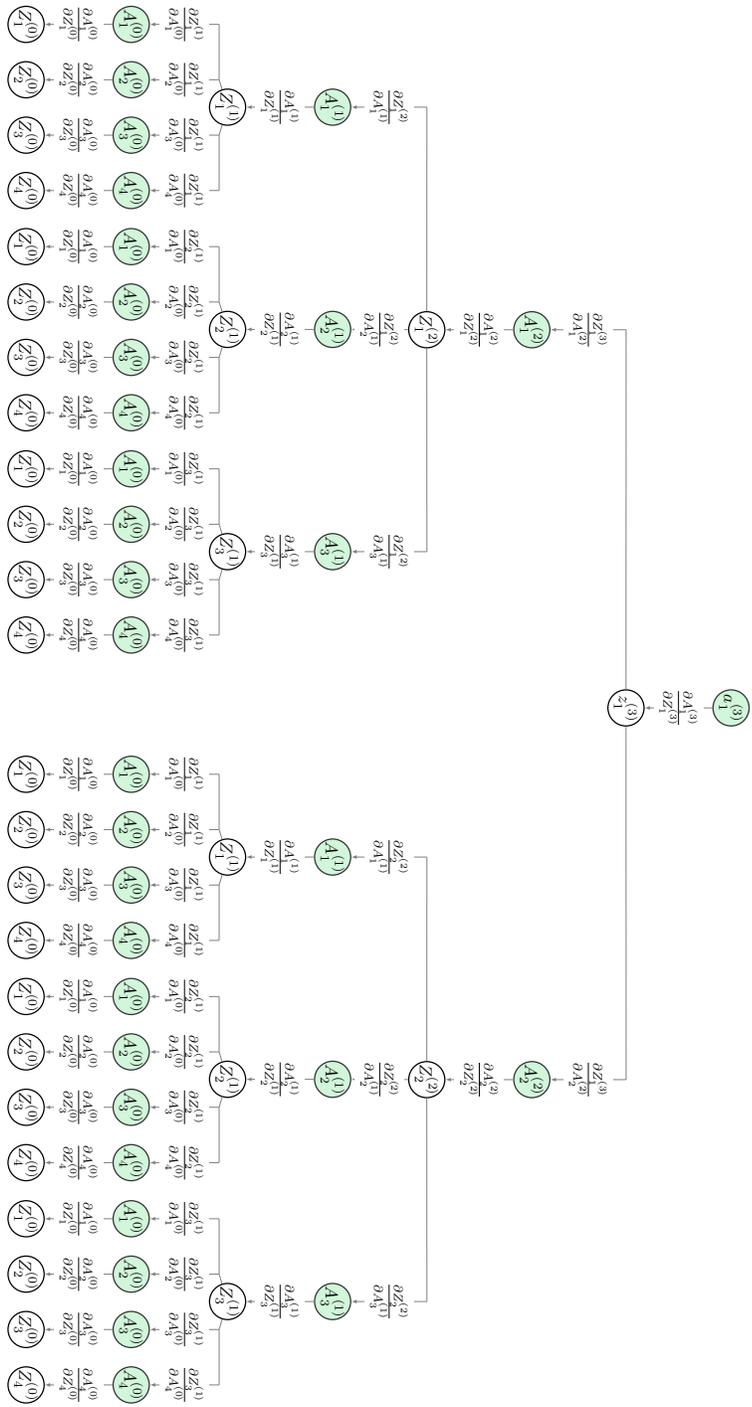}}}
\caption{Illustration of partial derivative diagram for a four-layer feedforward neural network.}
\label{fig_pdd}
\end{figure}

\begin{figure}[htbp]
\centering
\scalebox{0.57}{\rotatebox{-90}{\input{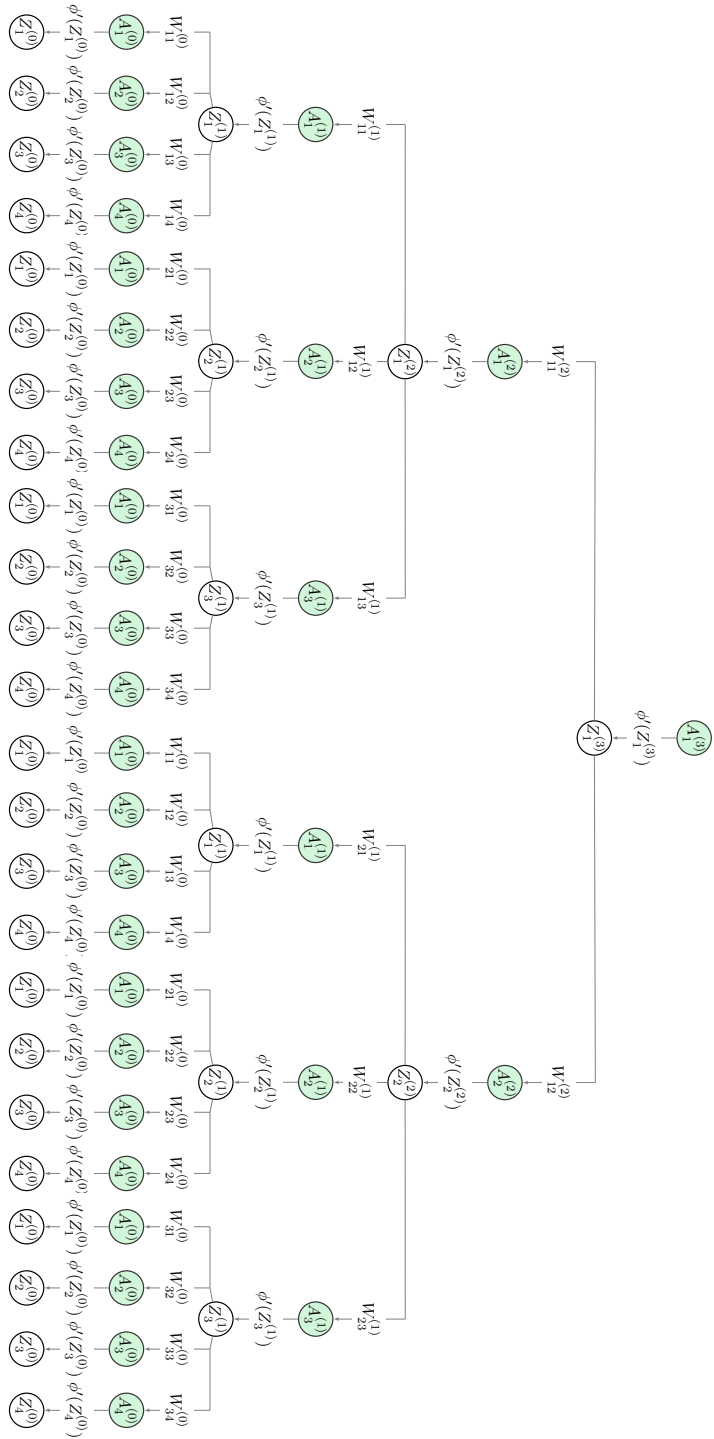}}}
\caption{Illustration of partial derivative diagram associated with the parameters and hidden states for a four-layer feedforward neural network.}
\label{fig_pddp}
\end{figure}

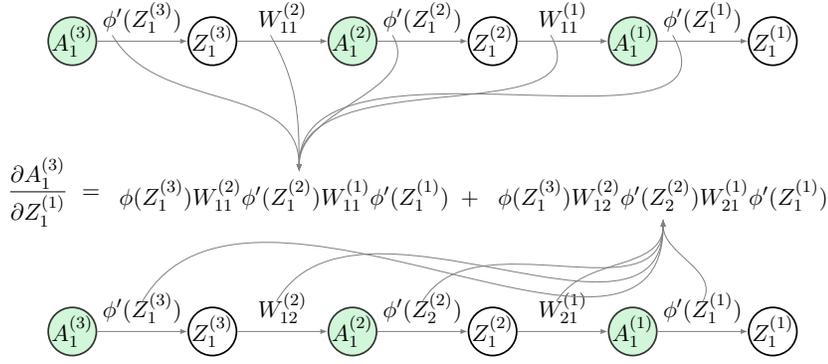
\begin{figure}[h!]
\begin{center}
\scalebox{0.8}{
\tikzstyle{connect}=[-latex, thick,>=latex]
\tikzstyle{shaded}=[draw=black!20,fill=black!20]
\tikzstyle{line} = [draw, -latex,thick,>=latex]

\tikzstyle{circ1}=[circle,minimum width=0.8cm,minimum height=0.8cm, thick, draw =black!100, inner ysep=-2pt,inner xsep=-4pt] 
\tikzstyle{circ2}=[circle, minimum width=0.8cm,minimum height=0.8cm,thick, draw =black!80,fill = green!80!blue!30, fill opacity=0.6,  text opacity=1,inner ysep=-2pt,inner xsep=-4pt] 

\begin{tikzpicture}
\node[circ2](a1)[]{$A_{1}^{(3)}$};
\node[circ1](z1)[right = 1.5cm of a1]{$Z_{1}^{(3)}$};
\node[circ2](a2)[right = 1.5cm of z1]{$A_{1}^{(2)}$};
\node[circ1](z2)[right = 1.5cm of a2]{$Z_{1}^{(2)}$};
\node[circ2](a3)[right = 1.5cm of z2]{$A_{1}^{(1)}$};
\node[circ1](z3)[right = 1.5cm of a3]{$Z_{1}^{(1)}$};
\node[circ2](a4)[below = 4cm of a1]{$A_{1}^{(3)}$};
\node[circ1](z4)[right = 1.5cm of a4]{$Z_{1}^{(3)}$};
\node[circ2](a5)[right = 1.5cm of z4]{$A_{1}^{(2)}$};
\node[circ1](z5)[right = 1.5cm of a5]{$Z_{1}^{(2)}$};
\node[circ2](a6)[right = 1.5cm of z5]{$A_{1}^{(1)}$};
\node[circ1](z6)[right = 1.5cm of a6]{$Z_{1}^{(1)}$};
\node(D)[below right = 1.5cm and -1.5cm of a1]{$\dfrac{\partial A_{1}^{(3)}}{\partial Z_{1}^{(1)}}\,\,=$};
\node(B1)[below right = -1.05cm and 0.05cm of D]{$\phi(Z_{1}^{(3)})W_{11}^{(2)}\phi'(Z_{1}^{(2)})W_{11}^{(1)}\phi'(Z_{1}^{(1)})\,\,\,+$};
\node(B2)[right = 0.05cm of B1]{$\phi(Z_{1}^{(3)})W_{12}^{(2)}\phi'(Z_{2}^{(2)})W_{21}^{(1)}\phi'(Z_{1}^{(1)})$};
\path (a1) edge[connect, draw=black!50, very thin]node[above]{$\phi'(Z_{1}^{(3)})$}(z1)
	 (z1) edge[connect, draw=black!50, very thin]node[above]{$W_{11}^{(2)}$}(a2)
	 (a2) edge[connect, draw=black!50, very thin]node[above]{$\phi'(Z_{1}^{(2)})$}(z2)
	 (z2) edge[connect, draw=black!50, very thin]node[above]{$W_{11}^{(1)}$}(a3)
	 (a3) edge[connect, draw=black!50, very thin]node[above]{$\phi'(Z_{1}^{(1)})$}(z3)
	 
	 (a4) edge[connect, draw=black!50, very thin]node[above]{$\phi'(Z_{1}^{(3)})$}(z4)
	 (z4) edge[connect, draw=black!50, very thin]node[above]{$W_{12}^{(2)}$}(a5)
	 (a5) edge[connect, draw=black!50, very thin]node[above]{$\phi'(Z_{2}^{(2)})$}(z5)
	 (z5) edge[connect, draw=black!50, very thin]node[above]{$W_{21}^{(1)}$}(a6)
	 (a6) edge[connect, draw=black!50, very thin]node[above]{$\phi'(Z_{1}^{(1)})$}(z6)
	 
	 ([xshift=2.5mm, yshift=1mm]a1.east) edge[connect, draw=black!50, very thin, out=-60, in = 90](B1)
	 ([xshift=5.5mm, yshift=1mm]z1.east) edge[connect, draw=black!50, very thin, out=-60, in = 90](B1)
	 ([xshift=2.5mm, yshift=1mm]a2.east) edge[connect, draw=black!50, very thin, out=-60, in = 90](B1)
	 ([xshift=5.5mm, yshift=1mm]z2.east) edge[connect, draw=black!50, very thin, out=-60, in = 90](B1)
	 ([xshift=2.5mm, yshift=1mm]a3.east) edge[connect, draw=black!50, very thin, out=-60, in = 90](B1)
	 
	 ([xshift=7.5mm, yshift=5mm]a4.east) edge[connect, draw=black!50, very thin, out=60, in = -90](B2)
	 ([xshift=6.5mm, yshift=5mm]z4.east) edge[connect, draw=black!50, very thin, out=60, in = -90](B2)
	 ([xshift=7.5mm, yshift=5mm]a5.east) edge[connect, draw=black!50, very thin, out=60, in = -90](B2)
	 ([xshift=6.5mm, yshift=5mm]z5.east) edge[connect, draw=black!50, very thin, out=60, in = -90](B2)
	 ([xshift=7.5mm, yshift=5mm]a6.east) edge[connect, draw=black!50, very thin, out=60, in = -90](B2)
;
\end{tikzpicture}}
\caption{Illustration of the partial derivative of $A_{1}^{(3)}$ with respect to $Z_{1}^{(1)}$.}
\label{fig_dl1}
\end{center}
\end{figure}

This section presents the partial derivative formulations for a feedforward neural network (FNN) of four layers in the context of TAGI. Figure \ref{fig_4lnn} presents the details of the interconnectivity of the variables associated with a four-layer FNN, 
Figure \ref{fig_pdd} describes the partial derivative diagram for the four-layer FNN presented in Figure \ref{fig_4lnn}, and
Figure \ref{fig_pddp} shows the partial derivative diagram associated with the parameters and hidden states. 
The partial derivative diagram allow computing the partial derivative of either a hidden state or an activation unit at any layers with respect to either the hidden state or activation unit from the previous layers. For example, the partial derivative of the first activation unit of layer three, i.e., $A_{1}^{(3)}$ with respect to the first hidden state of layer one, i.e., $Z_{1}^{(1)}$ is the sum of the product of the partial derivatives of  two branches relating to this partial derivative, which are identified using the partial derivative diagram in Figure \ref{fig_pddp}. Figure \ref{fig_dl1} illustrates the computations of this partial derivative.

\newpage
\subsubsection{Partial Derivative $\frac{\partial A_{1}^{(3)}}{\partial Z_{1}^{(2)}}$}
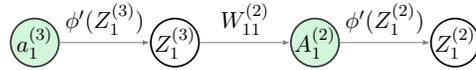
\begin{figure}[h!]
\begin{center}
\scalebox{0.8}{
\tikzstyle{connect}=[-latex, thick,>=latex]
\tikzstyle{shaded}=[draw=black!20,fill=black!20]
\tikzstyle{line} = [draw, -latex,thick,>=latex]

\tikzstyle{circ1}=[circle,minimum width=0.8cm,minimum height=0.8cm, thick, draw =black!100, inner ysep=-2pt,inner xsep=-4pt] 
\tikzstyle{circ2}=[circle, minimum width=0.8cm,minimum height=0.8cm,thick, draw =black!80,fill = green!80!blue!30, fill opacity=0.6,  text opacity=1,inner ysep=-2pt,inner xsep=-4pt] 

\begin{tikzpicture}
\node[circ2](a1)[]{$a_{1}^{(3)}$};
\node[circ1](z1)[right = 1.5cm of a1]{$Z_{1}^{(3)}$};
\node[circ2](a2)[right = 1.5cm of z1]{$A_{1}^{(2)}$};
\node[circ1](z2)[right = 1.5cm of a2]{$Z_{1}^{(2)}$};
\path (a1) edge[connect, draw=black!50, very thin]node[above]{$\phi'(Z_{1}^{(3)})$}(z1)
	 (z1) edge[connect, draw=black!50, very thin]node[above]{$W_{11}^{(2)}$}(a2)
	 (a2) edge[connect, draw=black!50, very thin]node[above]{$\phi'(Z_{1}^{(2)})$}(z2)
;
\end{tikzpicture}}
\caption{Illustration of a branch of the partial derivative of $A_{1}^{(3)}$ with respect to $Z_{1}^{(2)}$.}
\label{fig_l2_b1}
\end{center}
\end{figure}

This section presents the calculations of the partial derivative of $A_{1}^{(3)}$ with respect to $Z_{1}^{(2)}$. Figure \ref{fig_l2_b1} shows the branch from the partial derivative diagram (Figure \ref{fig_pddp}), that corresponds to this partial derivative.
This partial derivative is defined as
\begin{equation}
\dfrac{\partial A_{1}^{(3)}}{\partial Z_{1}^{(2)}} = \phi'(Z_{1}^{(3)})W_{11}^{(2)}\phi'(Z_{1}^{(2)}). 
\end{equation}
In the context of TAGI, the weights $W$ and hidden states $Z$ are Gaussian random variables, therefore, $\tfrac{\partial A_{1}^{(3)}}{\partial Z_{1}^{(2)}}$ is also approximated by a Gaussian PDF. The expected value is computed using Equation \ref{eq_m} and \ref{eq_idpApt},
\begin{equation}
\label{eq_mdl2}
\begin{array}{rcl}
\mathbb{E}\left[\dfrac{\partial A_{1}^{(3)}}{\partial Z_{1}^{(2)}} \right] &=&\mathbb{E}\left[\phi'(Z_{1}^{(3)})W_{11}^{(2)}\phi'(Z_{1}^{(2)})\right]\\[12pt]
 &=& \mathbb{E}\left[\phi'(Z_{1}^{(3)})\right]\,\mathbb{E}\left[W_{11}^{(2)}\phi'(Z_{1}^{(2)})\right] + \text{cov}\left(\phi'(Z_{1}^{(3)}), W_{11}^{(2)}\phi'(Z_{1}^{(2)})\right),
 \end{array}
\end{equation}
where
\begin{equation}
\label{eq_mdlw2}
\mathbb{E}\left[W_{11}^{(2)}\phi'(Z_{1}^{(2)})\right] = \mathbb{E}\left[W_{11}^{(2)}\right]\,\mathbb{E}\left[\phi'(Z_{1}^{(2)})\right] + \underbrace{\cancelto{0}{\text{cov}\left(W_{11}^{(2)}, \phi'(Z_{1}^{(2)})\right).}}_{\text{Eq.}\,\ref{eq_idpApt}}\\[12pt]
\end{equation}
\begin{equation}
\label{eq_covl2}
\begin{array}{rcl}
\text{cov}\left(\phi'(Z_{1}^{(3)}),\, W_{11}^{(2)}\phi'(Z_{1}^{(2)})\right)&=&\text{cov}\left(\phi'(Z_{1}^{(3)}),\, W_{11}^{(2)}\right)\,\mathbb{E}\left[\phi'(Z_{1}^{(2)})\right]\\[12pt]
&+&\text{cov}\left(\phi'(Z_{1}^{(3)}),\, \phi'(Z_{1}^{(2)})\right)\, \mathbb{E}\left[W_{11}^{(2)}\right].
\end{array}
\end{equation}
Note that the computations for the covariance $\text{cov}\left(\phi'(Z_{1}^{(3)}),\, W_{11}^{(2)}\right)$ and $\text{cov}\left(\phi'(Z_{1}^{(3)}),\, \phi'(Z_{1}^{(2)})\right)$ depend on the type of the activation function $\phi(.)$ being used for this layer (see \S \ref{SS:A:DA}). The variance is computed using Equation \ref{gma_var}, \ref{eq_idpApt}, \ref{eq_mdlw2}, and \ref{eq_mdl2},
\begin{equation}
\label{eq_vardl2}
\begin{array}{rcl}
\text{var}\left(\phi'(Z_{1}^{(3)})W_{11}^{(2)}\phi'(Z_{1}^{(2)})\right) &=& \text{var}\left(\phi'(Z_{1}^{(3)})\right)\,\text{var}\left(W_{11}^{(2)}\phi'(Z_{1}^{(2)})\right)\\[12pt] 
&+& \text{cov}\left(\phi'(Z_{1}^{(3)}), W_{11}^{(2)}\phi'(Z_{1}^{(2)})\right)^2\\[12pt]
&+& 2\text{cov}\left(\phi'(Z_{1}^{(3)}), W_{11}^{(2)}\phi'(Z_{1}^{(2)})\right)\,\\[12pt]
&\times&\mathbb{E}\left[W_{11}^{(2)}\right]\,\mathbb{E}\left[W_{11}^{(2)}\phi'(Z_{1}^{(2)})\right]\\[12pt]
&+&\text{var}\left(\phi'(Z_{1}^{(3)})\right)\, \mathbb{E}\left[W_{11}^{(2)}\phi'(Z_{1}^{(2)})\right]^2\\[12pt]
&+& \text{var}\left(W_{11}^{(2)}\phi'(Z_{1}^{(2)})\right)\, \mathbb{E}\left[\phi'(Z_{1}^{(3)})\right]^2,
\end{array}
\end{equation}
where
\begin{equation}
\begin{array}{rcl}
\text{var}\left(W_{11}^{(2)}\phi'(Z_{1}^{(2)})\right) &=& \text{var}\left(W_{11}^{(2)}\right)\,\text{var}\left(\phi'(Z_{1}^{(2)})\right) +  \underbrace{\cancelto{0}{\text{cov}\left(W_{11}^{(2)}, \phi'(Z_{1}^{(2)})\right)^2}}_{\text{Eq.}\,\ref{eq_idpApt}}\\[8pt]
&+&  2\cancelto{0}{\text{cov}\left(W_{11}^{(2)}, \phi'(Z_{1}^{(2)})\right)}\mathbb{E}\left[W_{11}^{(2)}\right]\,\mathbb{E}\left[\phi'(Z_{1}^{(2)})\right]\\[14pt]
&+& \text{var}\left(W_{11}^{(2)}\right)\,\mathbb{E}\left[\phi'(Z_{1}^{(2)})\right]^2 + \text{var}\left(\phi'(Z_{1}^{(2)})\right) \,\mathbb{E}\left[W_{11}^{(2)}\right]^2\\[14pt]
&=& \text{var}\left(W_{11}^{(2)}\right)\,\text{var}\left(\phi'(Z_{1}^{(2)})\right) +\text{var}\left(W_{11}^{(2)}\right)\,\mathbb{E}\left[\phi'(Z_{1}^{(2)})\right]^2\\[12pt] 
&+& \text{var}\left(\phi'(Z_{1}^{(2)})\right) \,\mathbb{E}\left[W_{11}^{(2)}\right]^2.
\end{array}
\end{equation}
\subsubsection{Partial Derivative $\frac{\partial A_{1}^{(3)}}{\partial Z_{1}^{(1)}} $}
\begin{figure}[htbp]
\begin{center}
\scalebox{0.8}{
\tikzstyle{connect}=[-latex, thick,>=latex]
\tikzstyle{shaded}=[draw=black!20,fill=black!20]
\tikzstyle{line} = [draw, -latex,thick,>=latex]

\tikzstyle{circ1}=[circle,minimum width=0.8cm,minimum height=0.8cm, thick, draw =black!100, inner ysep=-2pt,inner xsep=-4pt] 
\tikzstyle{circ2}=[circle, minimum width=0.8cm,minimum height=0.8cm,thick, draw =black!80,fill = green!80!blue!30, fill opacity=0.6,  text opacity=1,inner ysep=-2pt,inner xsep=-4pt] 

\begin{tikzpicture}
\node[circ2](a1)[]{$A_{1}^{(3)}$};
\node[circ1](z1)[right = 1.5cm of a1]{$Z_{1}^{(3)}$};
\node[circ2](a2)[right = 1.5cm of z1]{$A_{1}^{(2)}$};
\node[circ1](z2)[right = 1.5cm of a2]{$Z_{1}^{(2)}$};
\node[circ2](a3)[right = 1.5cm of z2]{$A_{1}^{(1)}$};
\node[circ1](z3)[right = 1.5cm of a3]{$Z_{1}^{(1)}$};
\path (a1) edge[connect, draw=black!50, very thin]node[above]{$\phi'(Z_{1}^{(3)})$}(z1)
	 (z1) edge[connect, draw=black!50, very thin]node[above]{$W_{11}^{(2)}$}(a2)
	 (a2) edge[connect, draw=black!50, very thin]node[above]{$\phi'(Z_{1}^{(2)})$}(z2)
	 (z2) edge[connect, draw=black!50, very thin]node[above]{$W_{11}^{(1)}$}(a3)
	 (a3) edge[connect, draw=black!50, very thin]node[above]{$\phi'(Z_{1}^{(1)})$}(z3)
;
\end{tikzpicture}}
\caption{Illustration of a branch of partial derivative of $A_{1}^{(3)}$ with respect to $Z_{1}^{(1)}$.}
\label{fig_l1_b1}
\end{center}
\end{figure}
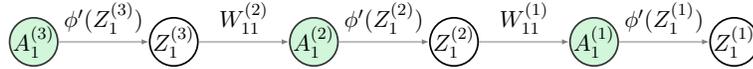

This section presents the calculations of the partial derivative of $A_{1}^{(3)}$ with respect to $Z_{1}^{(1)}$. According to the partial derivative diagram, there are two branches relating to this partial derivative. The partial derivative is a sum of  the product of partial derivatives on these two branches. The rest of this section only presents the computations for one of these two branches (Figure \ref{fig_l1_b1}).
This partial derivative is defined following
\begin{equation}
\dfrac{\partial A_{1}^{(3)}}{\partial Z_{1}^{(1)}} = \phi'(Z_{1}^{(3)})W_{11}^{(2)}\phi'(Z_{1}^{(2)})W_{11}^{(1)}\phi'(Z_{1}^{(1)}). 
\end{equation}
The expected value is computed using Equation \ref{eq_m}, \ref{gma_cov3}, \ref{eq_idpApt} and \ref{eq_mdl2},
\begin{equation}
\label{eq_mdl1}
\begin{array}{rcl}
\mathbb{E}\left[\dfrac{\partial A_{1}^{(3)}}{\partial Z_{1}^{(1)}}\right] &=& \mathbb{E}\left[\phi'(Z_{1}^{(3)})W_{11}^{(2)}\phi'(Z_{1}^{(2)})W_{11}^{(1)}\phi'(Z_{1}^{(1)})\right]\\[12pt]
&=& \mathbb{E}\left[\phi'(Z_{1}^{(3)})W_{11}^{(2)}\phi'(Z_{1}^{(2)})\right] \,\mathbb{E}\left[W_{11}^{(1)}\phi'(Z_{1}^{(1)})\right]\\[12pt]
&+& \text{cov}\left(\phi'(Z_{1}^{(3)})W_{11}^{(2)}\phi'(Z_{1}^{(2)}),\, W_{11}^{(1)}\phi'(Z_{1}^{(1)})\right),
\end{array}
\end{equation}
where
\begin{equation}
\label{eq_mdlw1}
\mathbb{E}\left[W_{11}^{(1)}\phi'(Z_{1}^{(1)})\right] = \mathbb{E}\left[W_{11}^{(1)}\right]\,\mathbb{E}\left[\phi'(Z_{1}^{(1)})\right] + \cancelto{0}{\text{cov}\left(W_{11}^{(1)}, \phi'(Z_{1}^{(1)})\right),}\\[12pt]
\end{equation}
\begin{equation}
\label{eq_covl1}
\begin{array}{rcl}
\text{cov}\left(\phi'(Z_{1}^{(3)})W_{11}^{(2)}\phi'(Z_{1}^{(2)}),\, W_{11}^{(1)}\phi'(Z_{1}^{(1)})\right)&&\\[4pt]
&\hspace{-10cm}=&\hspace{-5cm}\cancelto{0}{\text{cov}\left(\phi'(Z_{1}^{(3)}),\, W_{11}^{(1)}\phi'(Z_{1}^{(1)})\right)}\,\mathbb{E}\left[W_{11}^{(2)}\phi'(Z_{1}^{(2)})\right]\\[12pt]
&\hspace{-10cm}+&\hspace{-5cm}\text{cov}\left(W_{11}^{(2)}\phi'(Z_{1}^{(2)}),\, W_{11}^{(1)}\phi'(Z_{1}^{(1)})\right)\,\mathbb{E}\left[\phi'(Z_{1}^{(3)})\right]\\[8pt]
&\hspace{-10cm}=&\hspace{-5cm}\Bigg\{\text{cov}\left(\phi'(Z_{1}^{(2)}),\, W_{11}^{(1)}\phi'(Z_{1}^{(1)})\right)\,\mathbb{E}\left[W_{11}^{(2)}\right]\Bigg.\\[8pt]
&\hspace{-10cm}+&\hspace{-5cm}\Bigg. \cancelto{0}{\text{cov}\left(W_{11}^{(2)},\, W_{11}^{(1)}\phi'(Z_{1}^{(1)})\right)}\,\mathbb{E}\left[\phi'(Z_{1}^{(2)})\right]\Bigg\}\,\mathbb{E}\left[\phi'(Z_{1}^{(3)})\right]\\[12pt]
&\hspace{-10cm}=&\hspace{-5cm}\text{cov}\left(\phi'(Z_{1}^{(2)}),\, W_{11}^{(1)}\phi'(Z_{1}^{(1)})\right)\,\mathbb{E}\left[W_{11}^{(2)}\right]\,\mathbb{E}\left[\phi'(Z_{1}^{(3)})\right],\\[12pt]
\text{cov}\left(\phi'(Z_{1}^{(2)}),\, W_{11}^{(1)}\phi'(Z_{1}^{(1)})\right)\hspace*{2cm}&&\\[10pt]
&\hspace{-10cm}=&\hspace{-5cm}\text{cov}\left(\phi'(Z_{1}^{(2)}),\, W_{11}^{(1)}\right)\,\mathbb{E}\left[\phi'(Z_{1}^{(1)})\right]\\[12pt]
&\hspace{-10cm}+&\hspace{-5cm}\text{cov}\left(\phi'(Z_{1}^{(2)}),\, \phi'(Z_{1}^{(1)})\right)\, \mathbb{E}\left[W_{11}^{(1)}\right].
\end{array}
\end{equation}
The variance is computed using Equation \ref{gma_var}, \ref{eq_mdl2}, \ref{eq_vardl2}, \ref{eq_mdlw1},  and \ref{eq_covl1},
\begin{equation}
\label{eq_vardl1}
\begin{array}{rcl}
\text{var}\left(\dfrac{\partial A_{1}^{(3)}}{\partial Z_{1}^{(1)}}\right) &=& \text{var}\left(\phi'(Z_{1}^{(3)})W_{11}^{(2)}\phi'(Z_{1}^{(2)})W_{11}^{(1)}\phi'(Z_{1}^{(1)})\right)\\[12pt]
&=&\text{var}\left(\phi'(Z_{1}^{(3)})W_{11}^{(2)}\phi'(Z_{1}^{(2)})\right)\,\text{var}\left(W_{11}^{(1)}\phi'(Z_{1}^{(1)})\right)\\[12pt]
&+& \text{cov}\left(\phi'(Z_{1}^{(3)})W_{11}^{(2)}\phi'(Z_{1}^{(2)}),\,\, W_{11}^{(1)}\phi'(Z_{1}^{(1)})\right)^2\\[12pt]
&+&2\text{cov}\left(\phi'(Z_{1}^{(3)})W_{11}^{(2)}\phi'(Z_{1}^{(2)}),\,\, W_{11}^{(1)}\phi'(Z_{1}^{(1)})\right)\,\\[12pt]&&\mathbb{E}\left[\phi'(Z_{1}^{(3)})W_{11}^{(2)}\phi'(Z_{1}^{(2)})\right]\,\mathbb{E}\left[W_{11}^{(1)}\phi'(Z_{1}^{(1)})\right]\\[12pt]
&+&\text{var}\left(\phi'(Z_{1}^{(3)})W_{11}^{(2)}\phi'(Z_{1}^{(2)})\right)\,\mathbb{E}\left[W_{11}^{(1)}\phi'(Z_{1}^{(1)})\right]^2\\[12pt]
 &+& \text{var}\left(W_{11}^{(1)}\phi'(Z_{1}^{(1)})\right)\,\mathbb{E}\left[\phi'(Z_{1}^{(3)})W_{11}^{(2)}\phi'(Z_{1}^{(2)})\right]^2.
\end{array}
\end{equation}
The same above steps are repeated for the second branch in order to complete the calculation of the partial derivative of $A_{1}^{(3)}$ with respect to $Z_{1}^{(1)}$.

\subsubsection{Partial Derivative $\frac{\partial A_{1}^{(3)}}{\partial Z_{1}^{(0)}}$}\label{SSS:A:l0}
\begin{figure}[htbp]
\begin{center}
\scalebox{0.8}{
\tikzstyle{connect}=[-latex, thick,>=latex]
\tikzstyle{shaded}=[draw=black!20,fill=black!20]
\tikzstyle{line} = [draw, -latex,thick,>=latex]

\tikzstyle{circ1}=[circle,minimum width=0.8cm,minimum height=0.8cm, thick, draw =black!100, inner ysep=-2pt,inner xsep=-4pt] 
\tikzstyle{circ2}=[circle, minimum width=0.8cm,minimum height=0.8cm,thick, draw =black!80,fill = green!80!blue!30, fill opacity=0.6,  text opacity=1,inner ysep=-2pt,inner xsep=-4pt] 

\begin{tikzpicture}
\node[circ2](a1)[]{$A_{1}^{(3)}$};
\node[circ1](z1)[right = 1.2cm of a1]{$Z_{1}^{(3)}$};
\node[circ2](a2)[right = 1.2cm of z1]{$A_{1}^{(2)}$};
\node[circ1](z2)[right = 1.2cm of a2]{$Z_{1}^{(2)}$};
\node[circ2](a3)[right = 1.2cm of z2]{$A_{1}^{(1)}$};
\node[circ1](z3)[right = 1.2cm of a3]{$Z_{1}^{(1)}$};
\node[circ2](a4)[right = 1.2cm of z3]{$A_{1}^{(0)}$};
\node[circ1](z4)[right = 1.2cm of a4]{$Z_{1}^{(0)}$};
\path (a1) edge[connect, draw=black!50, very thin]node[above]{$\phi'(Z_{1}^{(3)})$}(z1)
	 (z1) edge[connect, draw=black!50, very thin]node[above]{$W_{11}^{(2)}$}(a2)
	 (a2) edge[connect, draw=black!50, very thin]node[above]{$\phi'(Z_{1}^{(2)})$}(z2)
	 (z2) edge[connect, draw=black!50, very thin]node[above]{$W_{11}^{(1)}$}(a3)
	 (a3) edge[connect, draw=black!50, very thin]node[above]{$\phi'(Z_{1}^{(1)})$}(z3)
	 (z3) edge[connect, draw=black!50, very thin]node[above]{$W_{11}^{(0)}$}(a4)
	 (a4) edge[connect, draw=black!50, very thin]node[above]{$\phi'(Z_{1}^{(0)})$}(z4);
\end{tikzpicture}}
\caption{Illustration of a branch of the partial derivative of $A_{1}^{(3)}$ with respect to $Z_{1}^{(0)}$.}
\label{fig_l0_B1}
\end{center}
\end{figure}
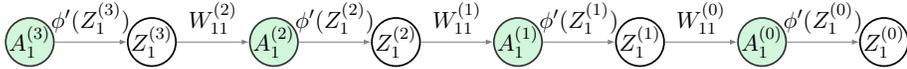

This section presents the calculations of the partial derivative of $A_{1}^{(3)}$ with respect to $Z_{1}^{(0)}$. From the partial derivative diagram (Figure \ref{fig_pddp}), we identify six branches relating to this partial derivative. Therefore, the partial derivative is equal to the sum of the product of partial derivatives on these six branches. Figure \ref{fig_l0_B1} shows the details for one of six branches. 
The partial derivative relating to this branch is defined following
\begin{equation}
\dfrac{\partial A_{1}^{(3)}}{\partial Z_{1}^{(0)}} = \phi'(Z_{1}^{(3)})W_{11}^{(2)}\phi'(Z_{1}^{(2)})W_{11}^{(1)}\phi'(Z_{1}^{(1)})W_{11}^{(0)}\phi'(Z_{1}^{(0)}).
\end{equation}
The expected value is computed using Equation \ref{eq_m}, \ref{eq_idpApt} and \ref{eq_mdl1},
\begin{equation}
\begin{array}{rcl}
\mathbb{E}\left[\dfrac{\partial A_{1}^{(3)}}{\partial Z_{1}^{(0)}}\right] &=& \mathbb{E}\left[\phi'(Z_{1}^{(3)})W_{11}^{(2)}\phi'(Z_{1}^{(2)})W_{11}^{(1)}\phi'(Z_{1}^{(1)})W_{11}^{(0)}\phi'(Z_{1}^{(0)})\right]\\[12pt]
&=&\mathbb{E}\left[\phi'(Z_{1}^{(3)})W_{11}^{(2)}\phi'(Z_{1}^{(2)})W_{11}^{(1)}\phi'(Z_{1}^{(1)})\right]\,\mathbb{E}\left[W_{11}^{(0)}\phi'(Z_{1}^{(0)})\right]\\[12pt]
&+&\text{cov}\left(\phi'(Z_{1}^{(3)})W_{11}^{(2)}\phi'(Z_{1}^{(2)})W_{11}^{(1)}\phi'(Z_{1}^{(1)}),\,\,W_{11}^{(0)}\phi'(Z_{1}^{(0)})\right),
\end{array}
\end{equation}
where
\begin{equation}
\label{eq_mdlw0}
\mathbb{E}\left[W_{11}^{(0)}\phi'(Z_{1}^{(0)})\right] = \mathbb{E}\left[W_{11}^{(0)}\right]\,\mathbb{E}\left[\phi'(Z_{0}^{(0)})\right] + \cancelto{0}{\text{cov}\left(W_{11}^{(1)},\, \phi'(Z_{1}^{(0)})\right),}
\end{equation}
\begin{equation}
\label{eq_covl0}
\begin{array}{rcl}
\text{cov}\left(\phi'(Z_{1}^{(3)})W_{11}^{(2)}\phi'(Z_{1}^{(2)})W_{11}^{(1)}\phi'(Z_{1}^{(1)}),\,\,W_{11}^{(0)}\phi'(Z_{1}^{(0)})\right) &&\\[0pt]
&\hspace{-10cm}=&\hspace{-5cm}\cancelto{0}{\text{cov}\left(\phi'(Z_{1}^{(3)})W_{11}^{(2)}\phi'(Z_{1}^{(2)}),\,\, W_{11}^{(0)}\phi'(Z_{1}^{(0)})\right)}\,\\[12pt]
&\hspace{-10cm}\times&\hspace{-5cm}\mathbb{E}\left[W_{11}^{(0)}\phi'(Z_{1}^{(0)})\right]\\[14pt]
&\hspace{-10cm}+&\hspace{-5cm}\text{cov}\left(W_{11}^{(1)}\phi'(Z_{1}^{(1)}),\,\,W_{11}^{(0)}\phi'(Z_{1}^{(0)})\right)\,\\[12pt]
&\hspace{-10cm}\times&\hspace{-5cm}\mathbb{E}\left[\phi'(Z_{1}^{(3)})W_{11}^{(2)}\phi'(Z_{1}^{(2)})\right]\\[12pt]
&\hspace{-10cm}=&\hspace{-5cm}\text{cov}\left(\phi'(Z_{1}^{(1)}),\,\,W_{11}^{(0)}\phi'(Z_{1}^{(0)})\right)\,\\[12pt]
&\hspace{-10cm}\times&\hspace{-5cm}\mathbb{E}\left[W_{11}^{(1)}\right]\,\underbrace{\mathbb{E}\left[\phi'(Z_{1}^{(3)})W_{11}^{(2)}\phi'(Z_{1}^{(2)})\right]}_{\text{Eq.}\,(\ref{eq_mdl2})}\\[22pt]
\text{cov}\left(\phi'(Z_{1}^{(1)}),\, W_{11}^{(0)}\phi'(Z_{1}^{(0)})\right)\hspace*{4cm}&&\\[8pt]
&\hspace{-10cm}=&\hspace{-5cm}\text{cov}\left(\phi'(Z_{1}^{(1)}),\, W_{11}^{(0)}\right)\,\mathbb{E}\left[\phi'(Z_{1}^{(0)})\right]\\[12pt]
&\hspace{-10cm}+&\hspace{-5cm}\text{cov}\left(\phi'(Z_{1}^{(1)}),\, \phi'(Z_{1}^{(0)})\right)\, \mathbb{E}\left[W_{11}^{(0)}\right].\\[12pt]
\end{array}
\end{equation}
The variance is computed using Equation \ref{gma_var}, \ref{eq_mdl1}, \ref{eq_vardl1}, \ref{eq_mdlw0} and \ref{eq_covl0},
\begin{equation}
\begin{array}{rcl}
\text{var}\left(\dfrac{\partial A_{1}^{(3)}}{\partial Z_{1}^{(0)}}\right) &=& \text{var}\left(\phi'(Z_{1}^{(3)})W_{11}^{(2)}\phi'(Z_{1}^{(2)})W_{11}^{(1)}\phi'(Z_{1}^{(1)})W_{11}^{(0)}\phi'(Z_{1}^{(0)})\right)\\[12pt]
&=&\text{var}\left(\phi'(Z_{1}^{(3)})W_{11}^{(2)}\phi'(Z_{1}^{(2)})W_{11}^{(1)}\phi'(Z_{1}^{(1)})\right)\,\text{var}\left(W_{11}^{(0)}\phi'(Z_{1}^{(0)})\right)\\[12pt]
 &+& \text{cov}\left(\phi'(Z_{1}^{(3)})W_{11}^{(2)}\phi'(Z_{1}^{(2)})W_{11}^{(1)}\phi'(Z_{1}^{(1)}),\,\,W_{11}^{(0)}\phi'(Z_{1}^{(0)})\right)^2\\[12pt]
 &+&2\text{cov}\left(\phi'(Z_{1}^{(3)})W_{11}^{(2)}\phi'(Z_{1}^{(2)})W_{11}^{(1)}\phi'(Z_{1}^{(1)}),\,\,W_{11}^{(0)}\phi'(Z_{1}^{(0)})\right)\\[12pt]
 &\times& \mathbb{E}\left[\phi'(Z_{1}^{(3)})W_{11}^{(2)}\phi'(Z_{1}^{(2)})W_{11}^{(1)}\phi'(Z_{1}^{(1)})\right]\,\mathbb{E}\left[W_{11}^{(0)}\phi'(Z_{1}^{(0)})\right]\\[12pt]
 &+&\text{var}\left(\phi'(Z_{1}^{(3)})W_{11}^{(2)}\phi'(Z_{1}^{(2)})W_{11}^{(1)}\phi'(Z_{1}^{(1)})\right)\,\mathbb{E}\left[W_{11}^{(0)}\phi'(Z_{1}^{(0)})\right]^2\\[12pt] 
 &+& \text{var}\left(W_{11}^{(0)}\phi'(Z_{1}^{(0)})\right)\,\mathbb{E}\left[\phi'(Z_{1}^{(3)})W_{11}^{(2)}\phi'(Z_{1}^{(2)})W_{11}^{(1)}\phi'(Z_{1}^{(1)})\right]^2.
\end{array}
\end{equation}
The same calculations are repeated for the five remaining branches in order to obtain the partial derivative of $A_{1}^{(3)}$ with respect to $Z_{1}^{(0)}$.
\subsubsection{Covariance between $\frac{\partial A_{1}^{(3)}}{\partial Z_{1}^{(0)}}$  and $Z_{1}^{(0)}$ }
This section presents the calculations of the covariance for the partial derivative $\tfrac{\partial A_{1}^{(3)}}{\partial Z_{1}^{(0)}}$ and $Z_{1}^{(0)}$. The following calculations correspond to the branch illustrated in Figure \ref{fig_l0_B1}, 
\begin{equation}
\begin{array}{rcl}
\text{cov}\left(\dfrac{\partial A_{1}^{(3)}}{\partial Z_{1}^{(0)}},\, Z_{1}^{(0)}\right) &&\\[16pt]
&\hspace{-5cm}=&\hspace{-2.5cm}\text{cov}\left(\phi'(Z_{1}^{(3)})W_{11}^{(2)}\phi'(Z_{1}^{(2)})W_{11}^{(1)}\phi'(Z_{1}^{(1)})W_{11}^{(0)}\phi'(Z_{1}^{(0)}),\,\,\,Z_{1}^{(0)}\right)\\[12pt]
&\hspace{-5cm}=&\hspace{-2.5cm}\text{cov}\left(\phi'(Z_{1}^{(3)})W_{11}^{(2)}\phi'(Z_{1}^{(2)})W_{11}^{(1)}\phi'(Z_{1}^{(1)}),\,\,\, Z_{1}^{(0)}\right)\,\underbrace{\mathbb{E}\left[W_{11}^{(0)}\phi'(Z_{1}^{(0)})\right]}_{\text{Eq.}\,(\ref{eq_mdlw0})}\\[22pt]
&\hspace{-5cm}+&\hspace{-2.5cm}\text{cov}\left(W_{11}^{(0)}\phi'(Z_{1}^{(0)}),\,\, Z_{1}^{(0)}\right)\,\underbrace{\mathbb{E}\left[\phi'(Z_{1}^{(3)})W_{11}^{(2)}\phi'(Z_{1}^{(2)})W_{11}^{(1)}\phi'(Z_{1}^{(1)})\right]}_{\text{Eq.}\,(\ref{eq_mdl1})}, 
\end{array}
\end{equation}
where
\begin{equation}
\begin{array}{rcl}
\text{cov}\left(\phi'(Z_{1}^{(3)})W_{11}^{(2)}\phi'(Z_{1}^{(2)})W_{11}^{(1)}\phi'(Z_{1}^{(1)}),\,\,\, Z_{1}^{(0)}\right) &&\\&\hspace{-14cm}=&\hspace{-7cm}\cancelto{0}{\text{cov}\left(\phi'(Z_{1}^{(3)})W_{11}^{(2)}\phi'(Z_{1}^{(2)}),\,\,Z_{1}^{(0)}\right)}\,\mathbb{E}\left[W_{11}^{(1)}\phi'(Z_{1}^{(1)})\right]\\[12pt]
&\hspace{-14cm}+&\hspace{-7cm}\text{cov}\left(W_{11}^{(1)}\phi'(Z_{1}^{(1)}),\,Z_{1}^{(0)}\right)\,\mathbb{E}\left[\phi'(Z_{1}^{(3)})W_{11}^{(2)}\phi'(Z_{1}^{(2)})\right]\\[12pt]
&\hspace{-14cm}=&\hspace{-7cm} \Bigg\{\cancelto{0}{\text{cov}\left(W_{11}^{(1)},\, Z_{1}^{(0)}\right)}\,\mathbb{E}\left[\phi'(Z_{1}^{(1)})\right]\Bigg.\\[12pt]
&\hspace{-14cm}+&\hspace{-7cm} \Bigg.\text{cov}\left(\phi'(Z_{1}^{(1)}),\, Z_{1}^{(0)}\right)\,\mathbb{E}\left[W_{11}^{(1)}\right]\Bigg\}\,\mathbb{E}\left[\phi'(Z_{1}^{(3)})W_{11}^{(2)}\phi'(Z_{1}^{(2)})\right]\\[12pt]
&\hspace{-14cm}=&\hspace{-7cm}\text{cov}\left(\phi'(Z_{1}^{(1)}),\, Z_{1}^{(0)}\right)\,\mathbb{E}\left[W_{11}^{(1)}\right]\,\underbrace{\mathbb{E}\left[\phi'(Z_{1}^{(3)})W_{11}^{(2)}\phi'(Z_{1}^{(2)})\right]}_{\text{Eq.}\,(\ref{eq_mdl2})}
\end{array}
\end{equation}
\begin{equation}
\begin{array}{rcl}
\text{cov}\left(W_{11}^{(0)}\phi'(Z_{1}^{(0)}),\,\, Z_{1}^{(0)}\right)&=& \cancelto{0}{\text{cov}\left(W_{11}^{(0)},\, Z_{1}^{(0)}\right)}\,\mathbb{E}\left[\phi'(Z_{1}^{(0)})\right]\\[12pt] 
&+&  \text{cov}\left(\phi'(Z_{1}^{(0)}),\, Z_{1}^{(0)}\right)\,\mathbb{E}\left[W_{11}^{(0)})\right].
\end{array}
\end{equation}
Note that the formulations for $\text{cov}\left(\phi'(Z_{1}^{(1)}),\, Z_{1}^{(0)}\right)$ and $\text{cov}\left(\phi'(Z_{1}^{(0)}),\, Z_{1}^{(0)}\right)$ are provided in \S \ref{SS:A:DA}. As mentioned in \S\ref{SSS:A:l0}, there are six branches relating to $\tfrac{\partial A_{1}^{(3)}}{\partial Z_{1}^{(0)}}$.  Therefore, we apply the same calculations for the five remaining branches. The final covariance between $\tfrac{\partial A_{1}^{(3)}}{\partial Z_{1}^{(0)}}$ and $Z_{1}^{(0)}$ is equal to the sum of the covariance of these branches.
\subsection{Generalization}
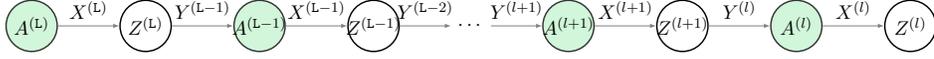
\begin{figure}[h!]
\begin{center}
\scalebox{0.68}{
\tikzstyle{connect}=[-latex, thick,>=latex]
\tikzstyle{shaded}=[draw=black!20,fill=black!20]
\tikzstyle{line} = [draw, -latex,thick,>=latex]

\tikzstyle{circ1}=[circle,minimum width=1cm,minimum height=1cm, thick, draw =black!100, inner ysep=-2pt,inner xsep=-4pt] 
\tikzstyle{circ2}=[circle, minimum width=1cm,minimum height=1cm,thick, draw =black!80,fill = green!80!blue!30, fill opacity=0.6,  text opacity=1,inner ysep=-2pt,inner xsep=-4pt] 

\begin{tikzpicture}
\node[circ2](a1)[]{$A^{(\mathtt{L})}$};
\node[circ1](z1)[right = 1.2cm of a1]{$Z^{(\mathtt{L})}$};
\node[circ2](a2)[right = 1.2cm of z1]{$A^{(\mathtt{L}-1)}$};
\node[circ1](z2)[right = 1.2cm of a2]{$Z^{(\mathtt{L}-1)}$};
\node(dd)[right = 1cm of z2]{\raisebox{0mm}{$\cdots$}};
\node[circ2](a3)[right = 1cm of dd]{$A^{(l+1)}$};
\node[circ1](z3)[right = 1.2cm of a3]{$Z^{(l+1)}$};
\node[circ2](a4)[right = 1.2cm of z3]{$A^{(l)}$};
\node[circ1](z4)[right = 1.2cm of a4]{$Z^{(l)}$};
\path (a1) edge[connect, draw=black!50, very thin]node[above]{$X^{(\mathtt{L})}$}(z1)
	 (z1) edge[connect, draw=black!50, very thin]node[above]{$Y^{(\mathtt{L}-1)}$}(a2)
	 (a2) edge[connect, draw=black!50, very thin]node[above]{$X^{(\mathtt{L}-1)}$}(z2)
	 (z2) edge[connect, draw=black!50, very thin]node[above]{$Y^{(\mathtt{L}-2)}$}(dd)
	 (dd) edge[connect, draw=black!50, very thin]node[above]{$Y^{(l+1)}$}(a3)
	 (a3) edge[connect, draw=black!50, very thin]node[above]{$X^{(l+1)}$}(z3)
	 (z3) edge[connect, draw=black!50, very thin]node[above]{$Y^{(l)}$}(a4)
	 (a4) edge[connect, draw=black!50, very thin]node[above]{$X^{(l)}$}(z4);
\end{tikzpicture}}
\caption{Illustration of a branch in the partial derivative diagram.}
\label{fig_ll_rb}
\end{center}
\end{figure}

This section presents the generalized formulations for a branch of the partial derivative diagram for a feedforward neural networks relating to the partial derivative of an activation unit at layer $\mathtt{L}$, i.e., $A^{(\mathtt{L})}$ with respect to a hidden state at layer $l$, i.e., $Z^{(l)}$. Figure \ref{fig_ll_rb} shows a branch of the partial derivative diagram for a FNN. 

\subsubsection{Partial Derivative}
\begin{equation}
\begin{array}{rcl}
\dfrac{\partial a^{(\mathtt{L})}}{\partial z^{(l)}} &=& X^{(\mathtt{L})}Y^{(\mathtt{L}-1)}X^{(\mathtt{L}-1)}.\cdots\,Y^{(l+1)}X^{(l+1)}Y^{(l)}X^{(l)}
\end{array}
\end{equation}
The expected value is computed following 
\begin{equation}
\scalebox{0.9}{$\begin{array}{rcl}
\mathbb{E}\left[\dfrac{\partial a^{(\mathtt{L})}}{\partial z^{(l)}} \right]&=& \underbrace{\mathbb{E}\left[X^{(\mathtt{L})}Y^{(\mathtt{L}-1)}X^{(\mathtt{L}-1)}\cdots\,Y^{(l+1)}X^{(l+1)}\right]}_{\mathbb{E}\left[\tfrac{\partial a^{(\mathtt{L})}}{\partial z^{(l+1)}} \right]}\,\mathbb{E}\left[Y^{(l)}\right]\mathbb{E}\left[X^{(l)}\right]\\[36pt]
&+&\text{cov}\left(X^{(\mathtt{L})}Y^{(\mathtt{L}-1)}X^{(\mathtt{L}-1)}\cdots\,Y^{(l+1)}X^{(l+1)},\,\,\, Y^{(l)}X^{(l)}\right), 
\end{array}$}
\end{equation}
where 
\begin{equation}
\scalebox{0.9}{$\begin{array}{rcl}
\text{cov}\left(X^{(\mathtt{L})}Y^{(\mathtt{L}-1)}X^{(\mathtt{L}-1)}\cdots\,Y^{(l+1)}X^{(l+1)},\,\,Y^{(l)}X^{(l)}\right) &&\\[8pt]
&\hspace{-10cm}=&\hspace{-5cm}\left\{\text{cov}\left(X^{(l+1)}, Y^{(l)}\right)\,\mathbb{E}\left[X^{(l)}\right]\right.\\[12pt]
&\hspace{-10cm}+&\hspace{-5cm}\left.\text{cov}\left(X^{(l+1)}, X^{(l)}\right)\,\mathbb{E}\left[Y^{(l)}\right]\right\}\\[12pt]
&\hspace{-10cm}\times&\hspace{-5cm}\mathbb{E}\left[Y^{(l+1)}\right]\,\underbrace{\mathbb{E}\left[X^{(\mathtt{L})}Y^{(\mathtt{L}-1)}X^{(\mathtt{L}-1)}\right]}_{\mathbb{E}\left[\tfrac{\partial a^{(\mathtt{L})}}{\partial z^{(\mathtt{L}-1)}}\right]}.
\end{array}$}
\end{equation}
The variance is computed following 
\begin{equation}
\scalebox{0.9}{$\begin{array}{rcl}
\text{var}\left(\dfrac{\partial a^{(\mathtt{L})}}{\partial z^{(l)}} \right)&=& \underbrace{\text{var}\left(X^{(\mathtt{L})}Y^{(\mathtt{L}-1)}X^{(\mathtt{L}-1)}\cdots\,Y^{(l+1)}X^{(l+1)}\right)}_{\text{var}\left(\tfrac{\partial a^{(\mathtt{L})}}{\partial z^{(l+1)}}\right)}\,\text{var}\left(Y^{(l)}X^{(l)}\right)\\[36pt]
&+& \text{cov}\left(X^{(\mathtt{L})}Y^{(\mathtt{L}-1)}X^{(\mathtt{L}-1)}\cdots\,Y^{(l+1)}X^{(l+1)},\,\,Y^{(l)}X^{(l)}\right)^2\\[12pt]
&+&2\text{cov}\left(X^{(\mathtt{L})}Y^{(\mathtt{L}-1)}X^{(\mathtt{L}-1)}\cdots\,Y^{(l+1)}X^{(l+1)},\,\,Y^{(l)}X^{(l)}\right)\\[12pt]
&\times&\underbrace{\mathbb{E}\left[X^{(\mathtt{L})}Y^{(\mathtt{L}-1)}X^{(\mathtt{L}-1)}\cdots\,Y^{(l+1)}X^{(l+1)}\right]}_{\mathbb{E}\left[\tfrac{\partial a^{(\mathtt{L})}}{\partial z^{(l+1)}} \right]}\,\mathbb{E}\left[Y^{(l)}\right]\mathbb{E}\,\left[X^{(l)}\right]\\[36pt]
&+&\text{var}\left(X^{(\mathtt{L})}Y^{(\mathtt{L}-1)}X^{(\mathtt{L}-1)}\cdots\,Y^{(l+1)}X^{(l+1)}\right)\,\mathbb{E}\left[Y^{(l)}\right]^2\,\mathbb{E}\left[X^{(l)}\right]^2\\[12pt]
&+&\text{var}\left(Y^{(l)}X^{(l)}\right)\,\mathbb{E}\left[X^{(\mathtt{L})}Y^{(\mathtt{L}-1)}X^{(\mathtt{L}-1)}\cdots\,Y^{(l+1)}X^{(l+1)}\right]^2.
\end{array}$}
\end{equation}

\subsubsection{Covariance between Partial Derivative and Hidden State}
\begin{equation}
\begin{array}{rcl}
\text{cov}\left(\dfrac{\partial a^{(\mathtt{L})}}{\partial z^{(l)}} ,\, z^{(l)}\right)&&\\[16pt]
&\hspace{-5cm}=&\hspace{-2.5cm}\text{cov}\left(X^{(l+1)},\,z^{(l)}\right)\,\mathbb{E}\left[Y^{(l+1)}\right]\,\mathbb{E}\left[Y^{(l)}\right]\mathbb{E}\,\left[X^{(l)}\right]\,\underbrace{\mathbb{E}\left[X^{(\mathtt{L})}Y^{(\mathtt{L}-1)}X^{(\mathtt{L}-1)}\right]}_{\mathbb{E}\left[\tfrac{\partial a^{(\mathtt{L})}}{\partial z^{(\mathtt{L}-1)}}\right]}\\[36pt]
&\hspace{-5cm}+&\hspace{-2.5cm}\text{cov}\left(X^{(l)},\, z^{(l)}\right)\,\mathbb{E}\left[Y^{(l)}\right]\, \underbrace{\mathbb{E}\left[X^{(\mathtt{L})}Y^{(\mathtt{L}-1)}X^{(\mathtt{L}-1)}\cdots\,Y^{(l+1)}X^{(l+1)}\right]}_{\mathbb{E}\left[\tfrac{\partial a^{(\mathtt{L})}}{\partial z^{(l+1)}} \right]}.
\end{array}
\end{equation}
\subsection{Activation Function}\label{SS:A:DA}
\subsubsection{Tanh(Z)}
The derivative of the function $\phi(Z) = tanh(Z)$ with respect to the hidden state $Z$ is written as 
\begin{equation}
\label{eq_dtanh}
\phi'(Z) = \dfrac{d \phi(z)}{dz} = 1 - \phi(Z)^2.
\end{equation}
The expected value of $\phi'(Z_{j}^{(l)}) $ is computed using Equation \ref{eq_m} and \ref{eq_dtanh}
\begin{equation}
\label{eq_dmtanh}
\begin{array}{rcl}
\mathbb{E}\left[\phi'(z)\right] &=& \mathbb{E}\left[1 - \phi(Z_{j}^{(l)})^2\right]\\[10pt]
&=& 1 - \mathbb{E}\left[\phi\left(Z_{j}^{(l)}\right)\right]^2 - \text{var}\left(\phi(Z_{j}^{(l)})\right).
\end{array}
\end{equation}
The variance of $\phi'(Z_{j}^{(l)}) $ is computed using Equation \ref{gma_var}
\begin{equation}
\label{eq_dvtanh}
\begin{array}{rcl}
\text{var}\left(\phi'(Z_{j}^{(l)})\right)&=& \text{var}\left(1 - \phi(Z_{j}^{(l)})^2\right)\\[10pt]
&=&\text{var}\left(\phi(Z_{j}^{(l)})^2\right)\\[10pt]
&=&2\,\text{var}\left(\phi(Z_{j}^{(l)})\right)\left\{\text{var}\left(Z_{j}^{(l)}\right) + 2\,\mathbb{E}\left[\phi(Z_{j}^{(l)})\right]^2\right\}.
\end{array}
\end{equation}
The covariance between $\phi(Z_{i}^{(l+1)})$ and $W_{ij}^{(l)}$ is computed using Equation \ref{gma_cov3}
\begin{equation}
\label{eq_dw}
\begin{array}{rcl}
\text{cov}\left(\phi'(Z_{i}^{(l+1)}),\, W_{ij}^{(l)}\right)&=&\text{cov}\left(1-\phi(Z_{i}^{(l+1)})^2,\, W_{ij}^{(l)}\right)\\[12pt]
&=&-\text{cov}\left(\phi(Z_{i}^{(l+1)})^2,\, W_{ij}^{(l)}\right)\\[12pt]
&=&-2\,\text{cov}\left(\phi(Z_{i}^{(l+1)}),\, W_{ij}^{(l)}\right)\,\mathbb{E}\left[\phi(Z_{i}^{(l+1)})\right].
\end{array}
\end{equation}
Using Equation \ref{eq_nnhs} and \ref{eq_lra}, Equation \ref{eq_dw} is rewritten as
\begin{equation}
\begin{array}{rcl}
\text{cov}\left(\phi'(Z_{i}^{(l+1)}),\, W_{ij}^{(l)}\right)&&\\[12pt]
&\hspace{-5cm}=&\hspace{-2.5cm}-2\,\text{cov}\left(J_{i}^{(l+1)}(Z_{i}^{(l+1)} - \mu_{Z_{i}}^{(l+1)}) + \phi(\mu_{Z_{i}}^{(l+1)}),\, W_{ij}^{(l)}\right)\,\mathbb{E}\left[\phi(Z_{i}^{(l+1)})\right]\\[10pt]
&\hspace{-5cm}=&\hspace{-2.5cm}-2\,J_{i}^{(l+1)}\text{cov}\left(Z_{i}^{(l+1)}, W_{ij}^{(l)}\right)\,\mathbb{E}\left[\phi(Z_{i}^{(l+1)})\right]\\[12pt]
&\hspace{-5cm}=&\hspace{-2.5cm}-2\,J_{i}^{(l+1)}\text{cov}\left(\displaystyle\sum_{k}W_{ik}^{(l)}\phi(Z_{k}^{(l)})+ B_{i}^{(l)},\,\, W_{ij}^{(l)}\right)\,\mathbb{E}\left[\phi(Z_{i}^{(l+1)})\right]\\[16pt]
&\hspace{-5cm}=&\hspace{-2.5cm}-2\,J_{i}^{(l+1)}\text{cov}\left(W_{ij}^{(l)}, W_{ij}^{(l)}\right)\,\mathbb{E}\left[\phi(Z_{i}^{(l)})\right]\,\mathbb{E}\left[\phi(Z_{i}^{(l+1)})\right].\\[10pt]
\end{array}
\end{equation}
The covariance between $\phi'(Z_{j}^{(l+1)})$ and $\phi'(Z_{i}^{(l)})$ is obtained using Equation \ref{gma_cov4}, \ref{eq_nnhs}, \ref{eq_idpApt}, and \ref{eq_lra},
\begin{equation}
\begin{array}{rcl}
\text{cov}\left(\phi'(Z_{i}^{(l+1)}),\, \phi'(Z_{j}^{(l)})\right) &=& \text{cov}\left(1-\phi(Z_{i}^{(l+1)})^2,\,\,1-\phi(Z_{j}^{(l)})^2\right)\\[12pt]
&=&\text{cov}\left(\phi(Z_{i}^{(l+1)})^2,\,\,\phi(Z_{j}^{(l)})^2\right)\\[12pt]
&=&2\,\text{cov}\left(\phi(Z_{i}^{(l+1)}),\,\, \phi(Z_{j}^{(l)})\right)^2 \\[12pt]
&+& 4\,\text{cov}\left(\phi(Z_{i}^{(l+1)}),\,\, \phi(Z_{j}^{(l)})\right)\,\mathbb{E}\left[\phi(Z_{i}^{(l+1)})\right]\,\mathbb{E}\left[\phi(Z_{j}^{(l)})\right],
\end{array}
\end{equation}
where
\begin{equation}
\begin{array}{rcl}
\text{cov}\left(\phi(Z_{i}^{(l+1)}),\,\, \phi(Z_{j}^{(l)})\right)&=&J_{i}^{(l+1)}\text{cov}\left(\displaystyle\sum_{k}W_{ik}^{(l)}\phi(Z_{k}^{(l)})+ B_{i}^{(l)},\,\,\phi(Z_{i}^{(l)})\right)\\[18pt]
&=&J_{i}^{(l+1)}\text{cov}\left(W_{ij}^{(l)}\phi(Z_{j}^{(l)}),\,\,\phi(Z_{i}^{(l)})\right)\\[12pt]
&=&J_{i}^{(l+1)}\text{cov}\left(\phi(Z_{j}^{(l)}),\,\,\phi(Z_{i}^{(l)})\right)\,\mathbb{E}\left[W_{ij}^{(l)}\right]\\[12pt]
&+& J_{i}^{(l+1)}\cancelto{0}{\text{cov}\left(W_{ij}^{(l)},\,\,\phi(Z_{i}^{(l)})\right)}\,\mathbb{E}\left[\phi(Z_{j}^{(l)})\right]. \\[12pt]

\end{array}
\end{equation}
The covariance between $\phi'(Z_{i}^{(l+1)})$ and $Z_{j}^{(l)}$ is computed using Equation \ref{gma_cov3}, \ref{eq_nnhs}, \ref{eq_idpApt}, and \ref{eq_lra},
\begin{equation}
\begin{array}{rcl}
\text{cov}\left(\phi'(Z_{i}^{(l+1)}),\,\, Z_{j}^{(l)}\right)&=&\text{cov}\left(1 - \phi(Z_{i}^{(l+1)})^2,\,\,Z_{j}^{(l)}\right)\\[12pt]
&=&-2\,\text{cov}\left(\phi(Z_{i}^{(l+1)}),\,\,Z_{j}^{(l)}\right)\,\mathbb{E}\left[Z_{j}^{(l)}\right]\\[12pt]
&=&-2\,J_{i}^{(l+1)}\text{cov}\left(\displaystyle\sum_{k}W_{ik}^{(l)}\phi(Z_{k}^{(l)})+ B_{i}^{(l)},\,\,Z_{j}^{(l)}\right)\,\mathbb{E}\left[Z_{j}^{(l)}\right]\\[18pt]
&=&-2\,J_{i}^{(l+1)}\text{cov}\left(W_{ij}^{(l)}\phi(Z_{j}^{(l)}),\,\,Z_{j}^{(l)}\right)\,\mathbb{E}\left[Z_{j}^{(l)}\right]\\[12pt]
&=&-2\,J_{i}^{(l+1)}\text{cov}\left(\phi(Z_{j}^{(l)}),\,\,Z_{j}^{(l)}\right)\,\mathbb{E}\left[W_{ij}^{(l)}\right]\,\mathbb{E}\left[Z_{j}^{(l)}\right]\\[12pt]
&-&2\,J_{i}^{(l+1)}\cancelto{0}{\text{cov}\left(W_{ij}^{(l)},\,\,Z_{j}^{(l)}\right)}\,\mathbb{E}\left[\phi(Z_{j}^{(l)})\right]\,\mathbb{E}\left[Z_{j}^{(l)}\right]\\[12pt]
&=&2\,J_{i}^{(l+1)}J_{j}^{(l)}\text{cov}\left(Z_{j}^{(l)},\,\,Z_{j}^{(l)}\right)\,\mathbb{E}\left[W_{ij}^{(l)}\right]\,\mathbb{E}\left[Z_{j}^{(l)}\right].
\end{array}
\end{equation}
The covariance between $\phi'(Z_{j}^{(l)})$ and $Z_{j}^{(l)}$ is computed using Equation \ref{gma_cov3}
\begin{equation}
\begin{array}{rcl}
\text{cov}\left(\phi'(Z_{j}^{(l)}),\,\,Z_{j}^{(l)}\right)&=&\text{cov}\left(1 - \phi(Z_{j}^{(l)})^2,\, Z_{j}^{(l)}\right)\\[12pt]
&=&-2\,\text{cov}\left(\phi(Z_{j}^{(l)}), Z_{j}^{(l)}\right)\,\mathbb{E}\left[\phi(Z_{j}^{(l)})\right]\\[12pt]
&=&-2\,J_{j}^{(l)}\text{cov}\left(Z_{j}^{(l)}, Z_{j}^{(l)}\right)\,\mathbb{E}\left[\phi(Z_{j}^{(l)})\right].
\end{array}
\end{equation}
\subsubsection{ReLU(Z)}
The derivative of the function $\phi(Z) = \text{ReLU}(Z)$ with respect to the hidden state $Z$ and its covariance are formulated following
\begin{equation}
\begin{array}{rcl}
\phi'(Z)&=&\left\{\begin{array}{rcl}
1&if& \mathbb{E}\left[Z\right]>0\\[4pt]
0 &if& \mathbb{E}\left[Z\right]\leq0.\end{array}\right.\\[14pt]
\mathbb{E}\left[\phi'(Z_{j}^{(l)})\right]&=&1\\[8pt]
\text{var}\left(\phi'(Z_{j}^{(l)})\right)&=&0\\[8pt]
\text{cov}\left(\phi'(Z_{i}^{(l+1)}),\, W_{ij}^{(l)}\right)&=&0\\[8pt]
\text{cov}\left(\phi'(Z_{i}^{(l+1)}),\, \phi'(Z_{j}^{(l)})\right)&=&0\\[8pt]
\text{cov}\left(\phi'(Z_{i}^{(l+1)}),\,\, Z_{j}^{(l)}\right)&=&0\\[8pt]
\text{cov}\left(\phi'(Z_{j}^{(l)}),\,\,Z_{j}^{(l)}\right)&=&0.
\end{array}
\end{equation}
\end{document}